%% file: main.tex
\documentclass[10pt,twocolumn,letterpaper]{article}

\usepackage[pagenumbers]{cvpr} % To force page numbers, e.g. for an arXiv version
\usepackage{graphicx}
\usepackage{amsmath}
\usepackage{amssymb}
\usepackage{booktabs}
\usepackage{multicol}
\usepackage{multirow}
\usepackage{overpic}

\usepackage{pifont}
\usepackage{nicefrac}       % compact symbols for 1/2, etc.
\usepackage{makecell}
\usepackage{microtype}   
\usepackage[accsupp]{axessibility}

\usepackage{algorithm}
\usepackage{algorithmic}
\usepackage{enumitem}

\input{preamble}

\definecolor{cvprblue}{rgb}{0.21,0.49,0.74}
\usepackage[pagebackref,breaklinks,colorlinks,citecolor=cvprblue]{hyperref}

\usepackage{nicefrac}
\usepackage{multirow}
\usepackage{booktabs}
\usepackage{float}
\usepackage{wrapfig}
\usepackage{bbding}
\usepackage{pifont}
\usepackage{color, colortbl}

\definecolor{gray}{gray}{0.9}
\newcommand{\ssymbol}[1]{^{\@fnsymbol{#1}}}

\def\paperID{11302} % *** Enter the Paper ID here
\def\confName{CVPR}
\def\confYear{2024}

\title{EfficientSAM: Leveraged Masked Image Pretraining for Efficient Segment Anything}

\author{Yunyang Xiong,\space Bala Varadarajan\thanks{Second author},\space Lemeng Wu\footnotemark[1],\space Xiaoyu Xiang,\space Fanyi Xiao,\space Chenchen Zhu,\\ Xiaoliang Dai,\space Dilin Wang,\space Fei Sun,\space Forrest Iandola,\space Raghuraman Krishnamoorthi,\space Vikas Chandra\\
Meta AI Research
}

\begin{document}
\maketitle
\input{sec/0_abstract}    
\input{sec/1_intro}

\input{sec/2_related}
\input{sec/3_method}

\input{sec/4_experiments}

\input{sec/5_conclusion}

{
    \small
    \bibliographystyle{ieeenat_fullname}
    \bibliography{main}
}

% WARNING: do not forget to delete the supplementary pages from your submission 
\input{sec/6_suppl}

\end{document}

%% file: preamble.tex
\usepackage[dvipsnames]{xcolor}

%% file: sec/0_abstract.tex
\begin{abstract}
Segment Anything Model (SAM) has emerged as a powerful tool for numerous vision applications. A key component that drives the impressive performance for zero-shot transfer and high versatility is a super large Transformer model trained on the extensive high-quality SA-1B dataset. While beneficial, the huge computation cost of SAM model has limited its applications to wider real-world applications. To address this limitation, we propose EfficientSAMs, light-weight SAM models that exhibits decent performance with largely reduced complexity. Our idea is based on leveraging masked image pretraining, SAMI, which learns to reconstruct features from SAM image encoder for effective visual representation learning. Further, we take SAMI-pretrained light-weight image encoders and mask decoder to build EfficientSAMs, and finetune the models on SA-1B for segment anything task. We perform evaluations on multiple vision tasks including image classification, object detection, instance segmentation, and semantic object detection, and find that our proposed pretraining method, SAMI, consistently outperforms other masked image pretraining methods. On segment anything task such as zero-shot instance segmentation, our EfficientSAMs with SAMI-pretrained lightweight image encoders perform favorably with a significant gain (e.g., $\sim $4 AP  on COCO/LVIS) over other fast SAM models. 
\end{abstract}

%% file: sec/1_intro.tex
\section{Introduction}
\label{sec:intro}
\begin{figure}[t]
    \centering
    \includegraphics[width=\linewidth]{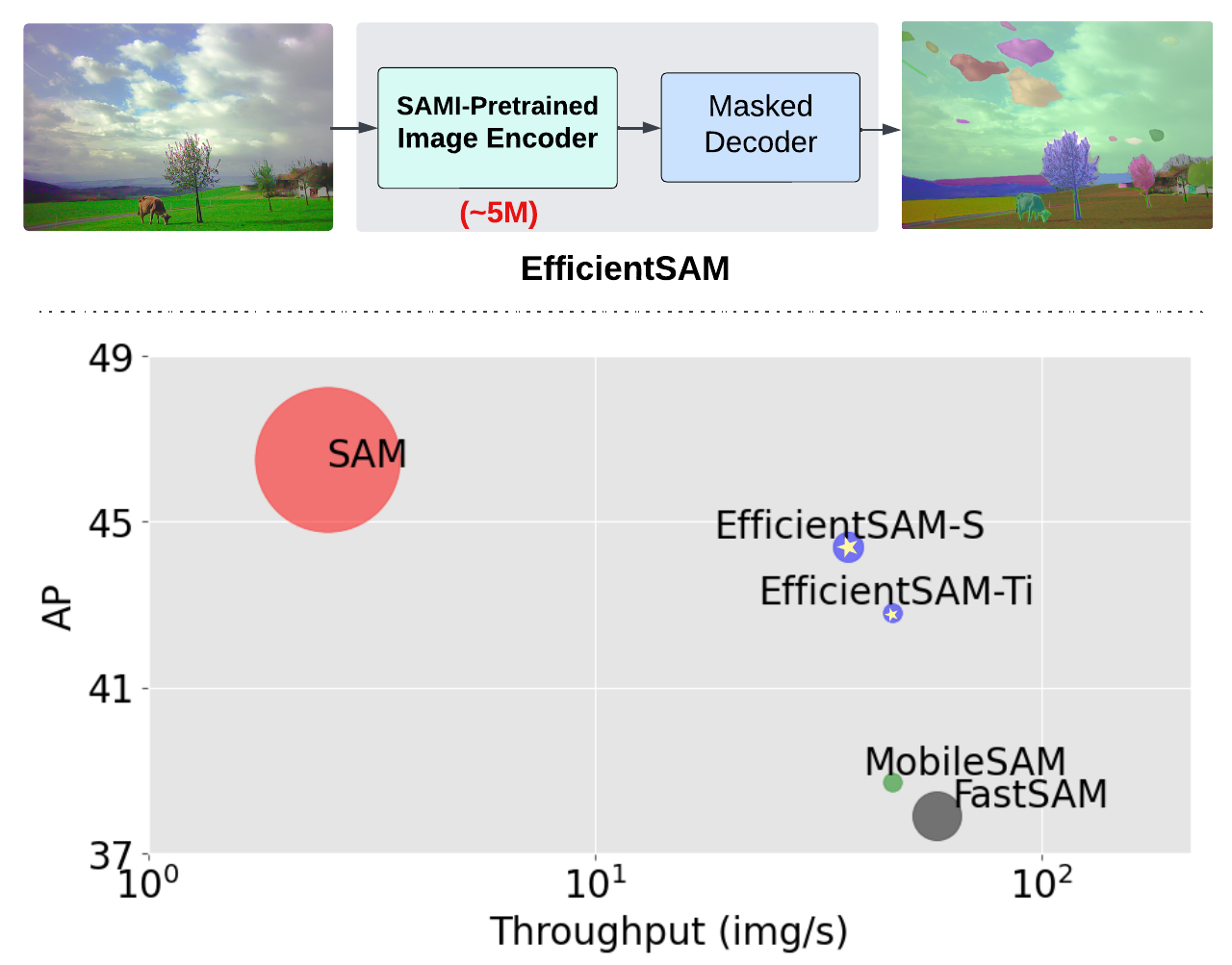}
    \caption{The comparative analysis result. (Top) The overview of EfficientSAM model by taking a well-pretrained light-weight image encoder for instance segmentation with largely reduced complexity. (Bottom) Throughput/Parameter/Performance comparison of EfficientSAM, MobileSAM, FastSAM, and SAM for zero-shot instance segmentation on COCO. We benchmark throughput (image per second) of all models on a single NVIDIA A100 with one box prompt. The input image resolution is $1024\times 1024$. Our EfficientSAMs outperform MobileSAM and FastSAM by a large margin, $\sim $4 AP, with comparable complexity. Our EfficientSAM-S reduces the inference time of SAM by $\sim $20x and the parameter size by $\sim $20x with a small performance drop, 44.4 AP vs 46.5 AP. }
    \label{fig:throughput}
\end{figure}

Segment Anything Model (SAM) \citep{kirillov2023segment} has been very successful in the vision field, achieving state-of-the-art performance in a variety of image segmentation tasks such as zero-shot edge detection\citep{arbelaez2010contour,kirillov2023segment}, zero-shot object proposal generation\citep{van2011segmentation,kirillov2023segment}, and zero-shot instance segmentation\citep{kirillov2023segment}, and many other real-world applications\citep{ma2023segment,tang2023can,han2023segment,sun2023explain,tariq2023segment,liu2023open}. The key feature of SAM is a prompt-based Vision Transformer (ViT)\citep{dosovitskiy2020image}model trained on a large-scale visual dataset with more than 1B masks from 11M images, SA-1B\citep{kirillov2023segment}, which allows segmenting any object on a given image. This ability of \textit{Segment Anything} makes SAM a foundation model in vision and enables its applications even beyond vision.

Despite the forgoing advantages, the model of SAM turns out to be a major efficiency bottleneck for practical deployment since the architecture of SAM, especially, the image encoder (e.g., ViT-H) is very expensive. Note that ViT-H image encoder in SAM has 632M parameters while the prompt-based decoder only take 3.87M parameters. As a result, it  leads to high computation and memory costs when using SAM to perform segment anything tasks in practice, which makes it challenging for real-time applications. 

To address this challenge, several recent works have proposed strategies that avoid incurring the huge cost when applying SAM for the prompt-based instance segmentation. For example, \citep{zhang2023faster} suggests distilling the knowledge from the default ViT-H image encoder to a tiny ViT image encoder. In \citep{zhao2023fast}, the computation cost can be reduced with a real-time CNN-based architecture for \textit{Segment Anything} task. 

In this paper, we propose using a well-pretrained lightweight ViT image encoder (e.g., ViT-Tiny/-Small\citep{touvron2021training}) to reduce the complexity of SAM while maintain decent performance. Our method, SAM-leveraged masked image pertraining (SAMI), produces desired pretrained lightweight ViT backbones for segment anything task. This is achieved by leveraging the celebrated MAE\citep{he2022masked} pretraining method with SAM model to obtain high-quality pretrained ViT encoders. Specifically, our SAMI makes use of SAM encoder, ViT-H, to generate feature embedding and train a masked image model with lightweight encoders to reconstruct features from ViT-H of SAM instead of image patches. This leads to generalized ViT backbones, which can be used for downstream tasks such as image classification, object detection, and segment anything. Then we finetune the pretrained lightweight encoders with SAM decoders for segment anything task\citep{kirillov2023segment}. 

To evaluate our method, we consider a transfer learning setting for masked image pretraining, where models are first pretrained with a reconstructive loss on ImageNet with image resolution $224\times 224$, and then finetuned on target tasks using supervised data. Our SAMI learns lightweight encoders that generalize well. With SAMI pretraining, we can train models like ViT-Tiny/-Small/-Base on ImageNet-1K with improved generalization performance. For a ViT-Small model, we achieve 82.7\% top-1 accuracy when finetuned on ImageNet-1K with 100 epochs, which outperforms other state-of-the-art image pretraining baselines. We also finetune our pretrained models on object detection, instance segmentation, and semantic segmentation. Across all these tasks, our pretraining method achieves better results than other pretraining baselines, and more importantly, we observe significant gains for small models. Further, we evaluate our models on Segment Anything task. On zero-shot instance segmentation, our model performs well compared to recent lightweight SAM methods, including FastSAM, by a margin of 4.1AP/5.2 AP on COCO/LVIS. 
% \begin{figure*}[!bhpt]
%     \centering
%     \includegraphics[width=0.8\textwidth]{figures/intro.pdf}
%     \caption{Caption}
%     \label{fig:intro}
% \end{figure*}

Our main contribution can be summarized as follows:

\begin{itemize}[noitemsep,topsep=0pt,leftmargin=0.75cm]
    \item 
    We propose a SAM-leveraged masked image pretraining framework  called \textbf{SAMI}, which trains the model to reconstruct features from SAM ViT-H image encoder. We show that this can substantially improve the performance of image masked pretraining method.

    \item We demonstrate that SAMI-pretrained backbones can generalize well to many tasks including image classification, object detection, and semantic segmentation.

    \item We deliver EfficientSAMs, light-weight SAM models with state-of-the-art quality-efficiency trade-offs (\cref{fig:throughput}), which is complementary to SAM for practical deployment. Code and models will be released to benefit a widge range of efficient SAM applications. 
    
\end{itemize}

%% file: sec/2_related.tex
\section{Related Work}
\label{sec:formatting}
We briefly review relevant works on segment anything model, vision transformers, knowledge distillation, and masked image pretraining. 

%-------------------------------------------------------------------------
\subsection{Segment Anything Model}
SAM\citep{kirillov2023segment} has been considered as a milestone vision foundation model, which can segment any object in the image based on interaction prompts. SAM has shown remarkable zero-shot transfer performance and high versatility for many vision tasks including a variety of segmentation application\citep{chen2023semantic,cen2023sad,deng2023segment,chen2023sam}, in-painting\citep{yu2023inpaint}, image restoration\citep{jiang2023restore}, image editing\citep{gao2023editanything}, image shadow removal\citep{zhang2023deshadow}, object tracking\citep{cheng2023segment,yang2023track}, and 3D object reconstruction\citep{shen2023anything}. There are many other works attempting to generalize SAM to real-world scenarios, including medical image segmentation\citep{ma2023segment}, camouflaged object detection\citep{tang2023can}, transparent object detection\citep{han2023segment}, concept-based explaination\citep{sun2023explain}, semantic communitation\citep{tariq2023segment}, and helping people with visual impairments\citep{liu2023open}. Due to its wide real-world applications, practical deployment of SAM has also gained increasing attention. Several recent works including \citep{zhang2023faster,zhao2023fast} have proposed strategies to reduce the computation costs of SAM. FastSAM\citep{zhang2023faster} develops a CNN-based architecture, YOLOv8-seg\citep{yolov8}, to segment all objects in an image for efficiency improvement. MobileSAM\citep{zhao2023fast} presents a decoupled distillation for obtaining a lightweight image encoder of SAM. Our work focuses on dealing with this efficiency issue for practical deployment of SAM.

%-------------------------------------------------------------------------
\subsection{Vision Transformers}
ViTs \citep{dosovitskiy2020image} have achieved impressive performance in vision applications\citep{carion2020end,fan2021multiscale,liu2021swin,li2022exploring,peng2022beit,he2022masked}. ViTs demonstrate the advantages of and generalization over their CNN counterparts\citep{he2022masked}. There are also a number of works on efficient ViTs for deployment. Smaller ViTs such as ViT-Small/Deit-Small and ViT-Tiny/DeiT-Tiny are introduced in \citep{touvron2021training} for complementing ViT-Huge, ViT-Large, and ViT-Base in \citep{dosovitskiy2020image}. Motivated by the ability of convolution to capture local information with reduced parameters and computation costs, MobileViT\citep{mehta2021mobilevit} explore combining ViT with convolutions, which outperforms light-weight CNN models such as MobileNet-v2/v3\citep{sandler2018mobilenetv2,koonce2021mobilenetv3} with better task-level generalization properties and reduced memory size and computation cost. This trick has been used in many follow-up works including LeViT\citep{graham2021levit},  EfficientFormer\citep{li2022efficientformer}, Next-ViT\citep{li2022next}, Tiny-ViT\citep{wu2022tinyvit}, Castling-ViT\citep{you2023castling}, EfficientViT\citep{liu2023efficientvit}. This line of progress for designing efficient ViTs is orthogonal to our EfficientSAM work towards building efficient SAM. 

%-------------------------------------------------------------------------
\subsection{Knowledge Distillation}
Knowledge distillation (KD) is a technique to improve the performance of deep learning models without changing their architectures. \citep{hinton2015distilling} is a pioneering work to distill the dark knowledge from a larger teacher model to a smaller student model. The learning of a student model is supervised by the hard labels and the soft labels from a teacher model. This practice is followed by multiple works which aim to make better use of soft labels to transfer more knowledge. In \citep{yang2021knowledge}, the distillation method decouples representation learning and classification. Decoupled knowledge distillation\citep{zhao2022decoupled} separates the classical KD loss into two parts, target class knowledge distillation and non-target class knowledge distillation, which improves the effectiveness and flexibility of knowledge transfer. Another line of work is to transfer knowledge from intermediate features. FitNet \citep{romero2014fitnets} is a pioneering work by distilling the semantic information from the teacher model's intermediate feature directly. In \citep{pmlr-v162-wu22c}, a self-supervised teaching assistant (SSTA) is introduced to guide the learning of a ViT-based student model with a supervised teacher together. \citep{bai2023masked} studies the potential of knowledge distillation from pre-training MAE model by aligning the intermediate features between the larger MAE teacher model and smaller MAE student model.

%-------------------------------------------------------------------------
\subsection{Masked Image Pretraining}
Self-supervised pretraining approaches \citep{caron2021emerging} have attracted significant attention in computer vision. One line of work is contrastive learning methods\citep{chen2020simple,chen2021exploring,xie2021propagate,wang2021dense}, which learn augmentation in-variance by imposing high similarity between different augmented views of  a given image. While the learned representation show good properties such as high linear separability, contrastive learning methods relies on strong augmentation and negative sampling. Another interesting line of work is masked image modeling (MIM), which helps models learn meaningful representations by reconstructing masked image patches. The MIM pioneering works focus on using denoising autoencoders\citep{vincent2010stacked} and context encoders\citep{pathak2016context} to train vision Transformer with masked prediction objectives. There are various promising works on using MIM for self-supervised image pretraining. BEiT\citep{bao2021beit} is the first one that adopts MIM for ViT pretraining to predict visual tokens. In BEiTv2\citep{peng2022beit}, a semantic-rich image tokenizer is utilized for a better reconstruction target. In MaskFeat\citep{wei2022masked}, reconstructing the local gradient features generated from HOG descriptor leads to effective visual pretraining. In SimMIM\citep{xie2022simmim} and MAE\citep{he2022masked}, directly reconstructing the pixel values of the masked image patches achieves effective visual representation learning. There are MAE-based follow-up works that use large teacher models to guide MAE pretraining\citep{pmlr-v162-wu22c,hou2022milan,bai2023masked}. Our work is built on MAE and finds that leveraging MAE to reconstruct the features from SAM image encoder enables the pretraining to be highly effective. 

%% file: sec/3_method.tex
\section{Approach}

\subsection{Preliminary}
\noindent \textbf{Masked Autoencoders.} Masked Autoencoders (MAE) model has two components, an encoder and a decoder. Both encoder and decoder are built on Transformer layers\citep{vaswani2017attention}. MAE takes image tokens, i.e., non-overlapping patches from the input images, as input. These input tokens are grouped to unmasked tokens and masked tokens with a given masking ratio. The unmasked tokens will be kept for encoder to extract features and the masked tokens will be set as the learning targets of the MAE decoder that need to be reconstructed during self-superivsed learning (MIM). MAE\citep{he2022masked} adopts a high mask ratio (e.g., 75\%), which prevents information leakage (e.g., simply extrapolating masked pixels based on the neighbors) in the pretraining stage. 

\begin{figure*}[!bhpt]
    \centering
    \includegraphics[width=0.95\textwidth]{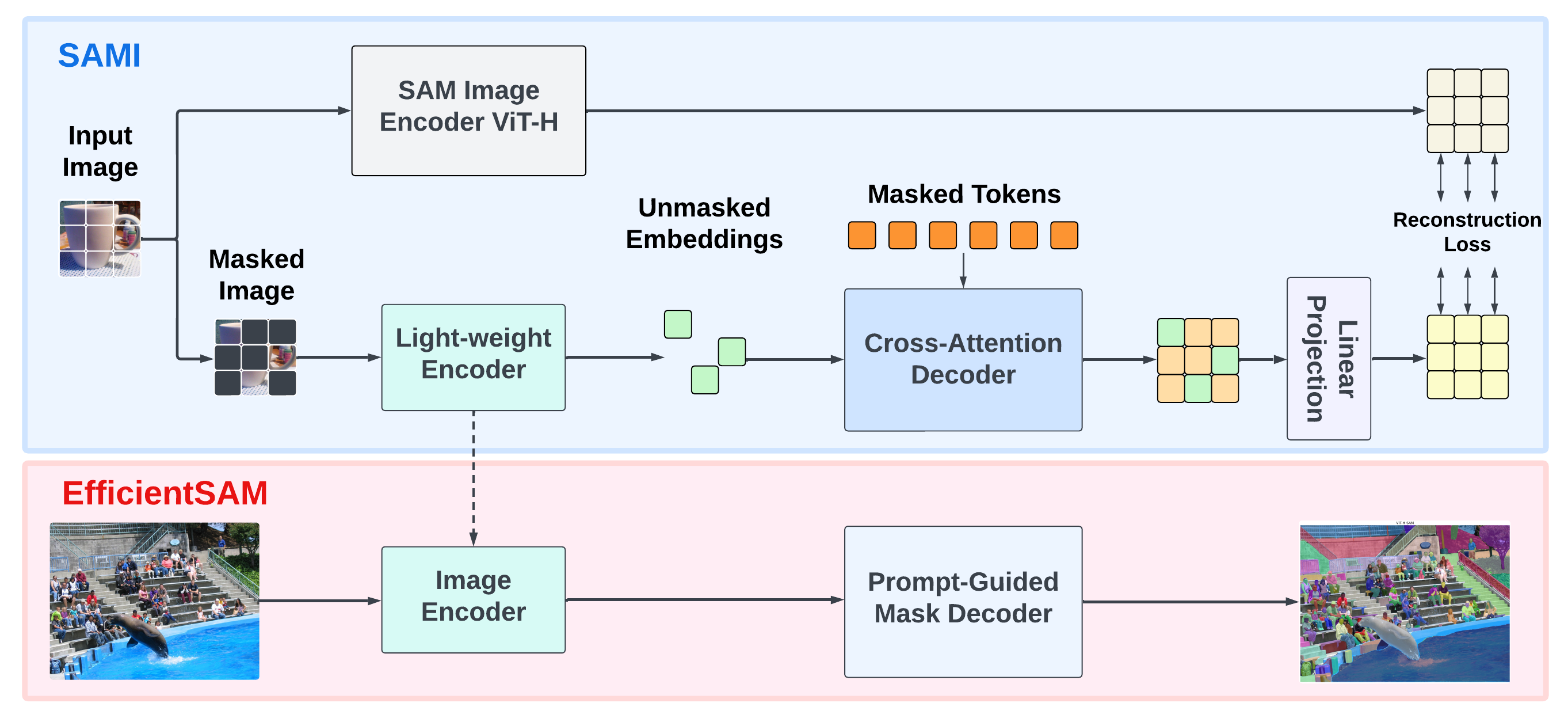}
    \caption{The overview of EfficientSAM framework. Our proposed EfficientSAM contains two stages: SAMI pretraining (top) on ImageNet and SAM finetuning (bottom) on SA-1B. For SAMI pretraining, the masked autoencoder takes the feature embeddings from SAM image encoder as the reconstruction target. After SAMI pretraining, the decoder is discarded and the light-weight encoder is served as the image encoder of EfficientSAM for finetuning on SA-1B. }
    \label{fig:efficientsams}
    \vspace{-3mm}
\end{figure*}

\subsection{SAM-Leveraged Masked Image Pretraining}
We now adapt MAE framework to obtain efficient image encoders for segment anything model. Motivated by the high versatility of SAM \citep{kirillov2023segment}, we explore latent features from SAM image encoder as the reconstruction target to leverage MAE. Our method emphasizes transferring the knowledge embedded in SAM. \cref{fig:efficientsams} (top) illustrates an overview of the proposed SAM-leveraged masked image pretraining, SAMI. The encoder transforms the unmasked tokens into latent feature representation and the decoder reconstructs the representation of the masked tokens aided by the output feature embedding from the encoder. The representation learning of reconstruction is guided by latent features from SAM. 

\textbf{Cross-Attention Decoder.} With the supervision of SAM features, we observe that only masked tokens need to be reconstructed via decoder while the encoder's output can serve as anchors during the reconstruction. In the cross-attention decoder, queries come from masked tokens, and keys and values derive from both unmasked features from encoder and masked features. We merge the output features of masked tokens from cross-attention decoder and the output features of unmasked tokens from encoder for the MAE output embedding. Then, this combined features will be reordered to the   original positions of input image tokens for the final MAE outputs. 

\textbf{Linear Projection Head.} We obtain the image output through our encoder and cross-attention decoder. Then we feed such features into a small project head for aligning the features from SAM image encoder. For simplicity, we just use a linear projection head to address the feature dimension mismatch between the output of SAM image encoder and MAE. 

\textbf{Reconstruction Loss.} At each training iteration, SAMI consists of a feedforward feature extraction from SAM image encoder, and a feedforward and a backpropagation procedure of MAE. The outputs from SAM image encoder and MAE linear projection head are compared to compute the reconstruction loss. 

Let us denote the SAM image encoder as $f^{\text{sam}}$, and the encoder and decoder of MAE  as $g^{e}$ with weights $W_e$ and $g^{d}$ with weights $W_d$, linear projection head as $h^{\theta}$ with weights $W_{\theta}$ respectively. Assume the input tokens are denoted as $\{\mathbf{x}_i\}_{i = 1}^N$, where $N$ is the number of tokens. The input tokens are randomly grouped into the unmasked tokens, $\{\mathbf{x}_i\}_{i \in \mathcal{U}}$, the masked tokens $\{\mathbf{x}_i\}_{i \in \mathcal{M}}$ with a given masked ratio. Let the feature reordering operator be $\phi$, and the merging operator be $\oplus$. 

Our target features from SAM image encoder can be written as $f^{\text{sam}}(\mathbf{x}) = f^{\text{sam}}(\{\mathbf{x}_i\}_{i = 1}^N)$, the output from MAE encoder is $g^{e}(\{\mathbf{x}_i\}_{i \in \mathcal{U}})$, the decoder output is $g^{d}(\{\mathbf{x}_i\}_{i \in \mathcal{M}})$. The output from linear projection head is $f^h(\mathbf{x}) = h^{\theta}(\phi(g^{e}\{\mathbf{x}_i\}_{i \in \mathcal{U}} \oplus g^{d}\{\mathbf{x}_i\}_{i \in \mathcal{M}}))$. Therefore, our target  reconstruction loss can be formulated as, 
\begin{equation} \label{equ:recon}
    \begin{split}
    L_{W_e, W_d, W_{\theta}} = \frac{1}{N} \cdot \sum_{j = 1}^N ||f^{\text{sam}}(\mathbf{x}) - f^h(\mathbf{x})||^2,
    \end{split}
\end{equation}
where $N$ is the number of input tokens, $||\cdot||$ denotes a norm. We use $\ell_2$ norm for reconstruction loss in our experiments. By minimizing the reconstruction loss, $L_{W_e, W_d, W_{\theta}}$, our encoder $g^{e}$ is optimized to serve as an image backbone to extract features as SAM image encoder. Our encoder, decoder, and linear projection head are optimized to learn context modeling ability from SAM image encoder. Optimizing the reconstruction loss on all tokens transfer the knowledge embedded in SAM.  

\textbf{SAMI for EfficientSAM.} After pretraining, our encoder extract feature representations for various vision tasks and the decoder is discarded. In particular, to build efficient SAM models for the segmentation anything task, we take the SAMI-pretrained lightweight encoder such as ViT-Tiny and ViT-Small as the image encoder and the default mask decoder of SAM for our EfficientSAM, as illustraed in \cref{fig:efficientsams} (bottom). We finetune our EfficientSAM models on SA-1B dataset for the segment anything task. The overview of our EfficientSAM framework is illustrated in \cref{fig:efficientsams}.

%% file: sec/4_experiments.tex
\section{Experiments}

\subsection{Experimental Settings}
\noindent \textbf{Pretraining Datasets.} Our masked image pretraining method, SAMI, is conducted on ImageNet-1K training set with 1.2M images. Following masked image pretraining \citep{he2022masked}, we do not use the label information. We use the SAM ViT-H image encoders from \citep{kirillov2023segment} to generate reconstruction features when pretraining our ViT models, ViT-Tiny, ViT-Small, and ViT-Base. 

\noindent \textbf{Pretraining Implementation Details.} Our ViT models are pretrained with a mean squared error (MSE) loss for reconstruction. We use a batch size of 4096, AdamW optimizer \citep{loshchilov2018decoupled} with learning rate 2.4e-3, $\beta_1 = 0.9$, $\beta_2 = 0.95$, weight decay 0.05, linear learning rate warm-up over the first $40$ epochs, cosine learning rate decay to update our models. We only adopt random resize crop to 224x224 resolution, random horiontal flip, and normalization for data augmentation. The mask ratio is set to $75\%$ and the decoder contains 8 Transformer blocks with 512 dimensions as in \citep{he2022masked}. We pretrain SAMI for 400 epochs using PyTorch framework on V100 machines. For reference, 1600-epoch pretraining is required for MAE\citep{he2022masked}. 

\noindent \textbf{Downstream Tasks/Datasets/Models.} \underline{\textit{Tasks and Datasets.}} We first consider three benchmarking datasets and several representative vision tasks to demonstrate the superiority of the proposed SAMI, including image classification on ImageNet dataset \cite{deng2009imagenet} with 1.2 million training and 50K validation images; 
Object detection and instance segmentation on COCO dataset \cite{lin2014microsoft} with 118K training and 5K validation images;
Semantic segmentation on ADE20K dataset \cite{zhou2017scene} with 20K/2K/3K images for training, validation, and testing, respectively.
Then, we consider segment anything task to further show the advantages of our proposed SAMI. We finetune our pretrained lightweight image encoders for SAM on SA-1B dataset \citep{kirillov2023segment} with more than 1B masks from 11M high-resolution images, and test interactive instance segmentation and zero-shot instance segmentation ability of our EfficientSAMs on COCO and LVIS \cite{gupta2019lvis}. 
\underline{\textit{Models.}} We discard the decoder of SAMI while keeping the encoder as backbone to extract features for different tasks as in MAE\citep{he2022masked}. We apply our well-pretrained ViT backbones for different tasks including ViTs for the classification, ViTDet \citep{li2022exploring} for the detection and instance segmentation, Mask2former \cite{cheng2022masked} for the semantic segmentation task, and SAM for segment anything.  

\noindent \textbf{Finetuning Settings.} \underline{\textit{For the classification task,}} we use the AdamW optimizer with $\beta_1 = 0.9$, $\beta_2 = 0.999$, weight decay $0.05$ to finetune ViTs for 100 epochs using 32 V100 GPUs, with each GPU having a batch size of 32.
The initial learning rate is 1e-3 with first $5$ epochs for linear warm-up and decays to zero by a cosine learning rate scheduler. We set the layer-wise learning rate decay factor to 0.75 for ViT-Small and ViT-Base. We do not apply layer-wise learning rate decay for ViT-Tiny. For data augmentation, we adopt RandAugment \citep{cubuk2020randaugment} and set label smoothing to 0.1, mixup to 0.8. 
\underline{\textit{For the detection and instance segmentation task,}} We follow the ViTDet \citep{li2022exploring} framework by adapting ViT backbones to a simple feature pyramid, for object detection and instance segmentation. We adopt AdamW optimizer with momentum $\beta_1 = 0.9$, $\beta_2 = 0.999$ and weight decay $0.1$ to train models on COCO.
All models are trained on 64 V100 GPUs for 100 epochs with each GPU having 1 batch size with image resolution $1024\times 1024$. The initial learning rate is $2e-4$, linearly warmed up for the first $10$ epochs, and decayed to 0 by a cosine learning rate schedule. Models are trained for 100 epochs.  
\underline{\textit{For the segmentation task,}} Our pretrained ViT models serve as the backbone of Mask2former \cite{cheng2022masked}, which is finetuned together with the segmentation layers on ADE20K. We adopt the AdamW optimizer with $\beta_1 = 0.9$, $\beta_2 = 0.999$, a mini-batch size of 16, a weight decay of 0.05, and an initial learning rate of 2e-4. The learning rate is decayed to 0 by a poly learning rate schedule. A learning rate multiplier is set to  0.1 for the backbone. The input image resolution is $512\times 512$. Models are trained for 160K iterations using 8 V100 GPUs. 
\underline{\textit{For the segmentation anything task,}} Following \citep{kirillov2023segment}, we take our pretrained lightweight ViT models, ViT-Tiny and ViT-Small, as the image encoder of SAM framework and finetune the encoder and decoder together of our EfficientSAM on SA-1B dataset for 5 epochs. We use the AdamW optimizer with a momentum, ($\beta_1 = 0.9$, $\beta_2 = 0.999$), a mini-batch size of 128, and a initial lrearning rate of $4e-4$. The learning rate is decayed to 0 by a linear learning rate schedule. We set weight decay to 0.1. We do not apply data augmentation. The input image resolution is $1024\times 1024$. Our EfficientSAMs are trained on 64 A100 GPUs with 40GB GPU memory. 

\noindent\textbf{Baselines and Evaluation Metrics.}
\underline{\textit{Baselines.}} For the classification task, we compare the performance of ViT backbones from different pretraining/distillation methods including MAE\citep{he2022masked}, 
SSTA\citep{pmlr-v162-wu22c},
DMAE\citep{bai2023masked},
BEiT\citep{bao2021beit}, CAE\citep{chen2023context}, DINO\citep{caron2021emerging}, iBOT\citep{zhou2021ibot}, DeiT\citep{touvron2021training}, etc.
For the detection and instance semantic task, and semantic segmentation task, we also compare with several pretrained ViT backbones for ViTDet\citep{li2022exploring} and Mask2former\citep{cheng2022masked}. For the segment everything task, we compare with SAM\citep{kirillov2023segment}, FastSAM\citep{zhao2023fast}, and MobileSAM\citep{zhang2023faster}. \underline{\textit{Evaluation Metrics.}} We evaluate our method and all baselines in terms of accuracy. Specifically, the accuracy metrics refer to top-1 accuracy for the classification task; AP$^{\text{box}}$, AP$^{\text{mask}}$, for the detection and  instance segmentation task (AP: average precision); mIoU, for the semantic segmentation task (mIoU: mean intersection over union); mIoU, AP, AP$^{\text{S}}$, AP$^{\text{M}}$, AP$^{\text{L}}$ for segment anything task. For efficiency metrics, we compare the number of model parameters or inference throughput.

\subsection{Main Results}
\input{tables/imagenet}
\textbf{Image Classification.} To evaluate the effectiveness of our proposed techniques on the image classification task, we apply the proposed SAMI idea to ViT models and compare their performance over baselines on ImageNet-1K.
As shown in \cref{tab:imagenet}, our SAMI is compared with pretraining methods like MAE, iBOT, CAE, and BEiT, and distillation methods including DeiT and SSTA. SAMI-B  achieves 84.8\% top-1 accuracy, which outperforms the pretraining baselines, MAE, DMAE, iBOT, CAE, and BEiT by 1.2\%, 0.8\%, 1.1\%, 0.9\%, and 0.4\% respectively. Compared with distillation methods such as DeiT and SSTA, SAMI also shows large improvements. For lightweight models such as ViT-Tiny and ViT-Small, SAMI reports a substantial gain compared to DeiT, SSTA, DMAE, and MAE. 

\input{tables/coco}
\noindent\textbf{Object Detection and Instance Segmentation.}
We also extend the SAMI-pretrained ViT backbones to the downstream object detection and instance segmentation task and compare it with previous pretraining baseline on COCO dataset to evaluate its efficacy. Specifically, we take the pretrained ViT backbones and adapt them to a simple feature pyramid in the Mask R-CNN framework\citep{he2017mask} for constructing the detector, ViTDet\citep{li2022exploring}. \cref{tab:coco} shows the overall comparison between our SAMI and other baselines. We can see that our SAMI consistently achieves better performance over other baselines. SAMI-B obtains 0.9 AP$^{\text{bbox}}$ and 0.6$^{\text{mask}}$ gains compared with MAE-B. For light-weight backbones, SAMI-S and SAMI-Ti report substantial gains compared to MAE-Ti and MAE-S. Moreover, SAMI-S significantly outperforms DeiT-S by 2.6 AP$^{\text{bbox}}$ and 2.3 AP$^{\text{mask}}$. For other pretraining baselines, our SAMI stiil compares favorably to DINO and iBOT. This set of experiments validate the effectiveness of the proposed SAMI for providing pretrained detector backbones in the object detection and instance segmentation task.  

\input{tables/ade20k}
\noindent\textbf{Semantic Segmentation.}
We further extend the pretrained backbones to the semantic segmentation task to evaluate its effectiveness.
Specifically, we use ViT models as the backbone in Mask2former~\cite{cheng2022masked} framework to benchmark on the ADE20K dataset.
As shown in Tab. \ref{tab:ade20k}, Mask2former with SAMI-pretrained  backbones achieve better mIoU, i.e., $\uparrow$2.5,  $\uparrow$4.7, and $\uparrow$3.7 improvement over backbones with MAE pretraining~\cite{he2022masked} on ImageNet-1K. This set of experiments validate that our proposed techniques could be well generalized to various downstream tasks.

\subsection{EfficientSAMs for Segment Anything Task}
Segment Anything task is a process of promptable segmentation to produce segmentation masks based on any form of the prompt, including point set, rough boxes or mask, free-form text. We follow SAM \citep{kirillov2023segment} and focus on  point-based and box-based prompt segmentation on COCO/LVIS. We now test the generalization abilities of our model on segment anything task including zero-shot single point valid mask evaluation and zero-shot instance segmentation. We take the SAMI-pretrained lightweight backbones as the image encoder of SAM for building efficient SAMs, EfficientSAMs. Then we finetune EfficientSAMs on SA-1B dataset and report the performance on zero-shot single point valid mask evaluation and zero-shot instance segmentation. 

\input{tables/sam_zero_shot_point}
\textbf{Zero-Shot Single Point Valid Mask Evaluation.} Similar to SAM\citep{kirillov2023segment}, we evaluate segmenting an object from a single foreground point. For general interactive segmentation, we also consider object segmentation from a single box, and multiple points as introduced in \citep{kirillov2023segment}. To achieve this, we uniformly sample random points within ground truth mask for click, and compute the tightest bounding box corresponding to ground truth mask for box. Since our models are able to predict multiple masks, we only evaluate the most confident mask as SAM \citep{kirillov2023segment}. 

\textbf{Results.} In \cref{tab:sam0point}, EfficientSAMs are compared with SAM, MobileSAM and SAM-MAE-Ti. On COCO, our EfficientSAM-Ti outperforms MobileSAM by 1.9 mIoU on 1 click and 1.5 mIoU on 1 box with comparable complexity. Our EfficientSAM-Ti with SAMI-pretrained weights also performs better than MAE-pretrained wights on COCO/LVIS interactive segmentation. We notice that our EfficientSAM-S only underperforms SAM by 1.5 mIoU on COCO box and 3.5 mIoU on LVIS box with 20x fewer parameters. We find that our EfficientSAMs also show promising performance on multiple click compared with MobileSAM and SAM-MAE-Ti. 

\input{tables/sam_zero_shot_inseg}
\textbf{Zero-Shot Instance Segmentation.} Following SAM \citep{kirillov2023segment}, instance segmentation task is performed by taking the bounding box (bbox) generated by ViTDet\citep{li2022exploring} as the prompt. The mask with the highest Intersection over Union (IoU) with the bbox as the predicted mask. 

\textbf{Results.} In \cref{tab:sam0inseg}, we report AP, AP$^{\text{S}}$, AP$^{\text{M}}$, AP$^{\text{L}}$ for zero-shot instance segmentation. We compare our EfficientSAM with MobileSAM and FastSAM. We can see that EfficientSAM-S obtains more than 6.5 AP on COCO and 7.8 AP on LVIS over FastSAM. For EffidientSAM-Ti, it still outperforms FastSAM by a large margin, 4.1 AP on COCO and 5.3 AP on LVIS, and MobileSAM by 3.6 AP on COCO and 5.5 AP on LVIS. Note that our EfficientSAMs are much light-weight than FastSAM, e.g, 9.8M parameters for efficientSAM-Ti vs 68M parameters for FastSAM. EfficientSAM-S also significantly reduces the gap between SAM with 0.6G parameters, only $\sim$2 AP reduction. These results demonstrate the extraordinary benefits of EfficientSAMs for zero-shot instance segmentation and validate the advantages of our SAMI pretraining method. 

\begin{figure}
    \centering
    \begin{overpic}[width=1.0\linewidth]{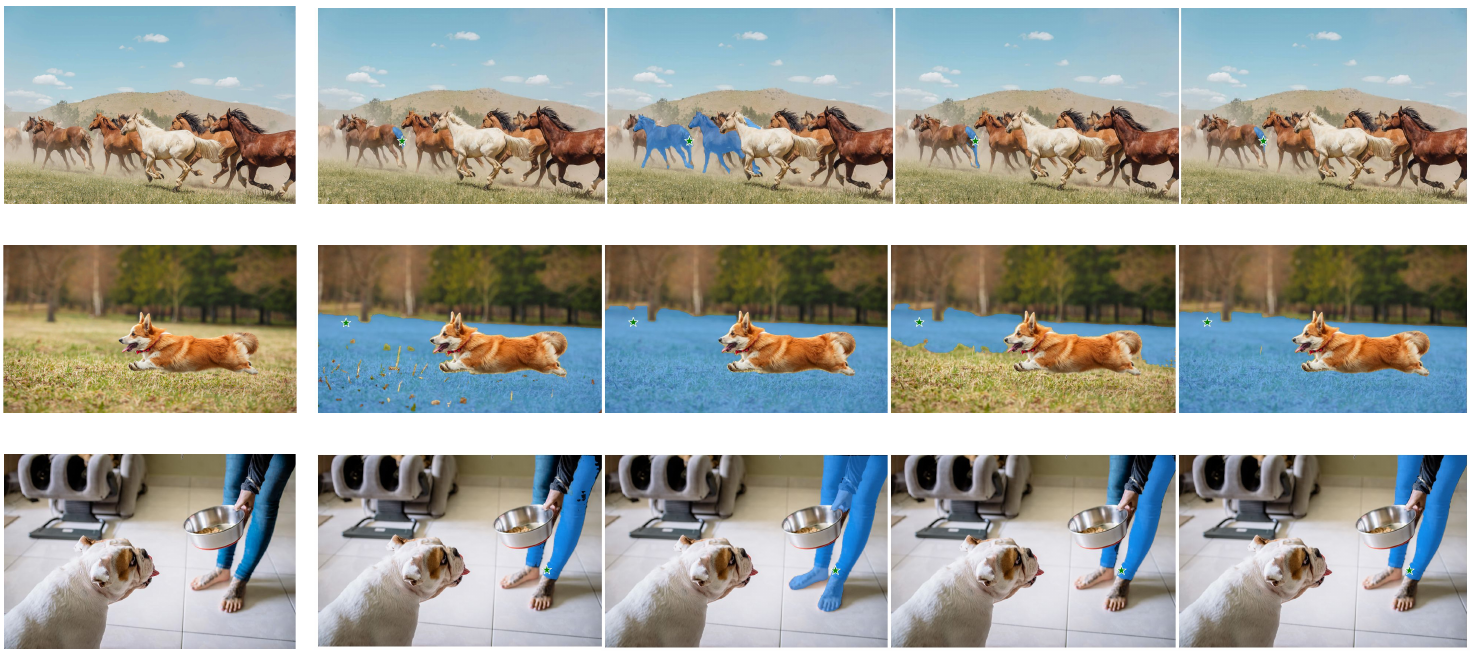}
\put (2,46) {\scriptsize{Input Image}}
\put (25,46) {\scriptsize{SAM\citep{kirillov2023segment}}}
\put (42,46) {\scriptsize{FastSAM\citep{zhao2023fast}}}
\put (60,46) {\scriptsize{MobileSAM\citep{zhang2023faster}}}
\put (82,46) {\scriptsize{EfficientSAM}}
\end{overpic}
    \caption{Visualization results on point-prompt input with SAM, FastSAM, MobileSAM,  and our EfficientSAM model.}
    \label{fig:point_prompt}
\end{figure}

\begin{figure}[h]
\centering
\vspace{5pt}
\begin{overpic}[width=\linewidth]{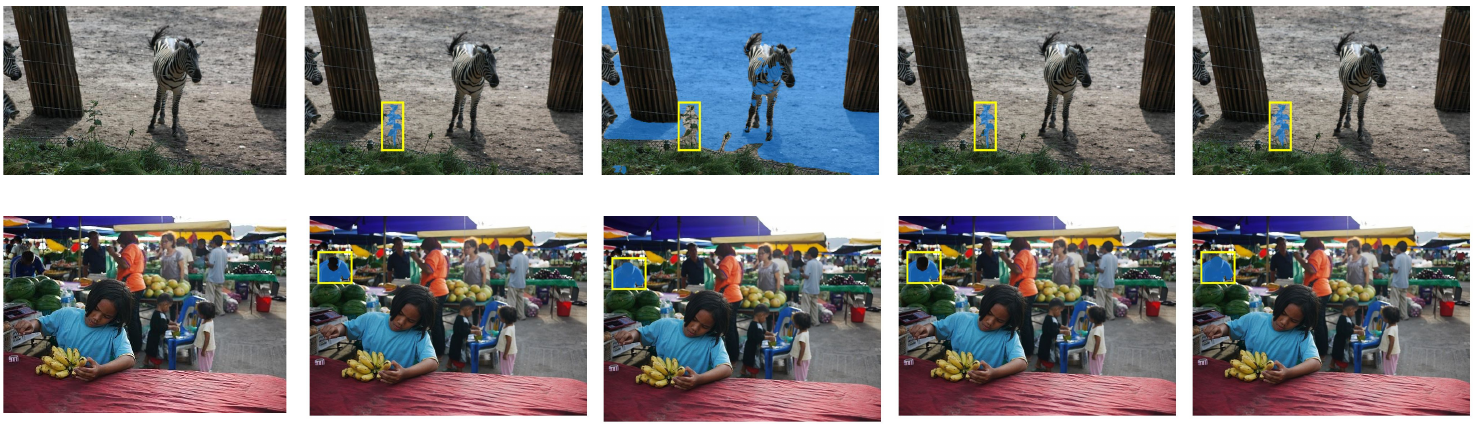}
\put (2,30) {\scriptsize{Input Image}}
\put (25,30) {\scriptsize{SAM\citep{kirillov2023segment}}}
\put (42,30) {\scriptsize{FastSAM\citep{zhao2023fast}}}
\put (60,30) {\scriptsize{MobileSAM\citep{zhang2023faster}}}
\put (82,30) {\scriptsize{EfficientSAM}}
\end{overpic}
\caption{Visualization results on box-prompt input with SAM, FastSAM, MobileSAM, and our EfficientSAM  model.}
\label{fig:box_prompt}
\end{figure}

\begin{figure}
    \centering
    \begin{overpic}[width=1.0\linewidth]{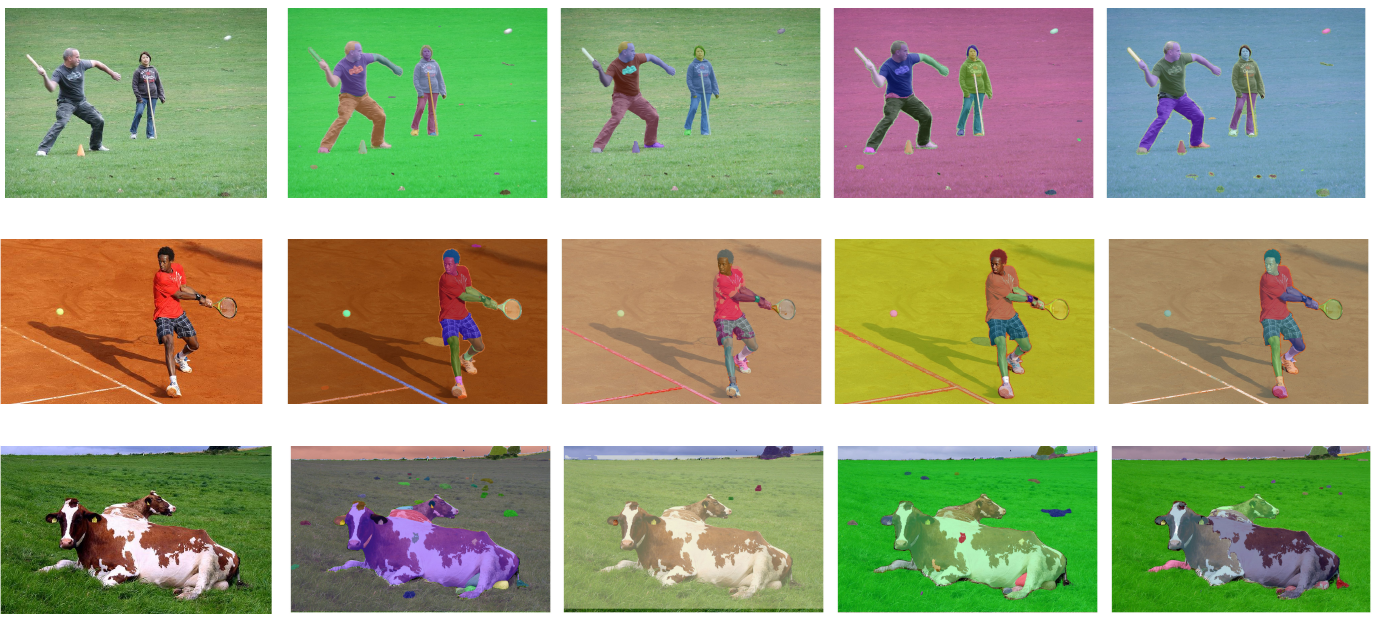}
\put (2,46) {\scriptsize{Input Image}}
\put (25,46) {\scriptsize{SAM\citep{kirillov2023segment}}}
\put (42,46) {\scriptsize{FastSAM\citep{zhao2023fast}}}
\put (60,46) {\scriptsize{MobileSAM\citep{zhang2023faster}}}
\put (82,46) {\scriptsize{EfficientSAM}}
\end{overpic}
    \caption{Visualization results on segment everything with SAM, FastSAM, MobileSAM, and our EfficientSAM model.}
    \label{fig:everything_prompt}
\end{figure}
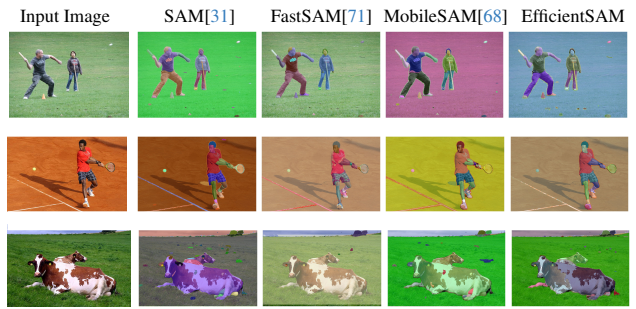

\textbf{Qualitative Evaluation.} We now provide the qualitative results for a complementary understanding of instance segmentation capabilities of EfficientSAMs. Some examples can be seen in  \cref{fig:point_prompt}, \cref{fig:box_prompt}, and \cref{fig:everything_prompt}. Specifically, we report the predicted masks with two types of prompts, point and box as in MobileSAM \citep{zhang2023faster} and also segment everything results. More qualitative results can be found in the supplement. These results demonstrate that our EfficientSAMs have competing capabilities when comparing to SAM. Note that our EfficientSAMs are much lightweight than SAM, and our models can effectively give decent segmentation results. This indicates that our models can be served as a complementary version of SAM for many practical tasks. 

\begin{figure}[!bpht]
    \centering
    \vspace{5pt}
    \begin{overpic}[width=0.8\linewidth]{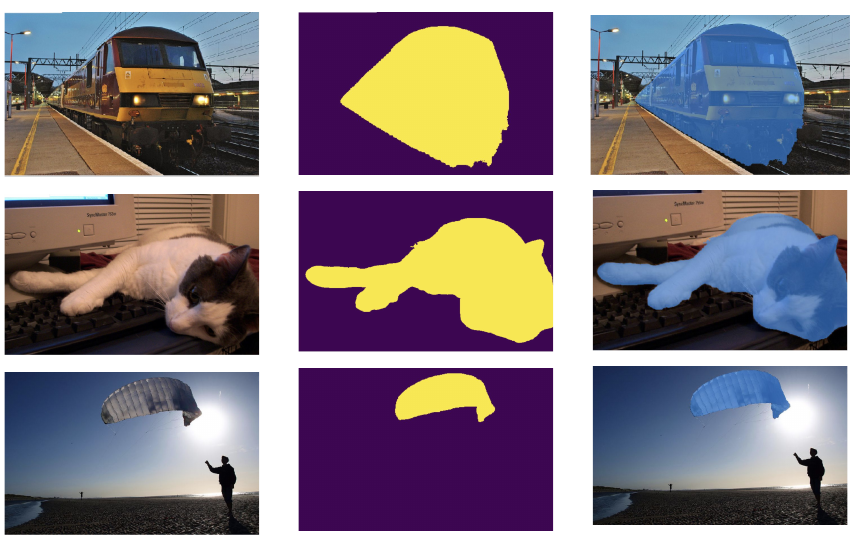}
\put (6,64) {\scriptsize{Input Image}}
\put (46,64) {\scriptsize{Mask}}
\put (74,64) {\scriptsize{Segmentation}}
\end{overpic}
    \caption{Saliency-based automatic instance segmentation results. With the assistance of saliency map generated from U$^2$-net\citep{Qin_2020_PR}, our EfficientSAM is able to generate mask and perform automatic instance segmentation without manually creating points or boxes. }
    \label{fig:saliency}
\end{figure}

\textbf{Salient Instance Segmentation.} Salient object segmentation \citep{borji2019salient} aims to segment the most visually attractive objects from an image. We extend interactive instance segmentation to salient instance segmentation without manually creating points/boxes. Specifically, we take a state-of-the-art saliency object detection model, U$^2$-net\citep{Qin_2020_PR}, to predict saliency map and uniformly sample 3 random points (3 click) within saliency map to perform instance segmentation with our EfficientSAM. In \cref{fig:saliency}, we can see that our EfficientSAM can perform salient instance segmentation well. This preliminary exploration  opens the potential to help people with hand impairments segment objects of interest in an image.

\subsection{Ablation Studies}
We now analyze SAMI and EfficientSAMs through a series of ablation studies with ViT backbones. 

\input{tables/loss}
\textbf{Reconstruction Loss.} We study the effect of reconstruction loss on the performance of SAMI on ImageNet-1K. We compare our mean square error (MSE) reconstruction loss with cosine similarity loss. We find that MSE reconstruction loss performs better, shown in \cref{tab:loss}. This recommends a direct reconstruction of SAM features instead of the target with a high angular similarity.  

\textbf{Cross-Attention Decoder.} To reconstruct SAM features, we directly use the output tokens from encoder and only take decoder to transform the masked tokens with cross-attention. We study how the performance varies with all tokens through decoder as MAE\citep{he2022masked}. When querying the masked tokens in decoder, we find that SAMI-Ti performs 3\% better on ImageNet-1K  than feeding all tokens into decoder for target reconstruction as MAE\citep{he2022masked}. Analogy to anchor points in AnchorDETR\citep{wang2022anchor}, the output tokens from encoder are already learned well by directly aligning the SAM features, which can serve as anchor tokens for help masked tokens align via cross-attention decoder.  

\input{tables/mask_ratio}
\textbf{Mask Ratio.} A high mask ratio, 75\%, is recommended in MAE\citep{he2022masked}. We explore how the performance varies with different mask ratio in SAMI. As shown in \cref{tab:maskratio}, we can see the observations are consistent with MAE \citep{he2022masked} that a high mask ratio tends to produce good results. 

\textbf{Reconstruction Target.} We study the impact of reconstruction target. We take a different encoder from CLIP \citep{ramesh2021zero} to generate features as the reconstruction target in SAMI. Aligning features from CLIP encoder can also outperform MAE by 0.8\% for a ViT-Tiny model on ImageNet-1K. This demonstrates that masked image pretraining benefits from powerful guided reconstruction. 
\begin{figure}[!bpth]
    \centering
    \includegraphics[width=0.85\linewidth]{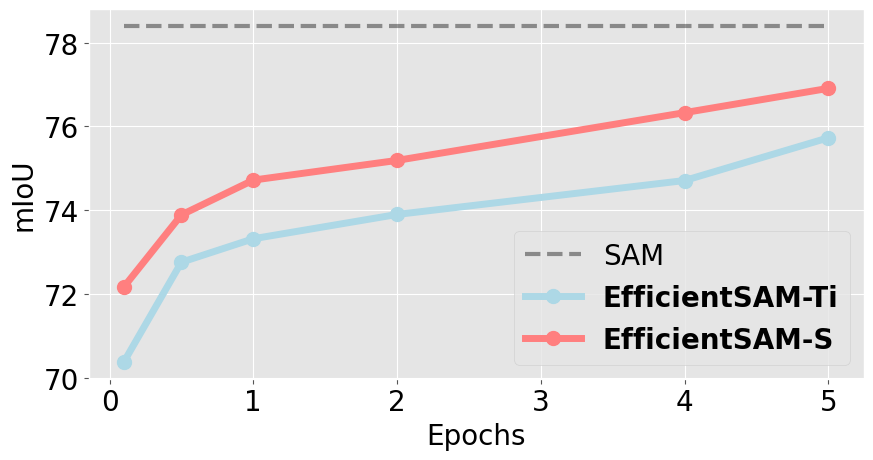}
    \caption{Ablation study on training steps for EfficientSAMs on MS COCO dataset. Zero-shot single point valid mask evaluation with a single box prompt is performed for the ablation.}
    \label{fig:ablation_step}
\end{figure}

\textbf{Effects of Finetuning Steps for EfficientSAMs.} We explore the effect of fintuning steps for EfficientSAMs. As illustrated in \cref{fig:ablation_step}, EfficientSAM-Ti and EfficientSAM-S achieve decent performance even at $0.1$ epoch. For 1 epoch, the performance gain is larger than 2.5 mIoU. The final performance of EfficientSAM-S reaches 76.9 mIoU, which is only 1.5 mIoU lower than SAM. These results demonstrate the advantages of SAMI-pretrained image encoders and our EfficientSAMs.

%% file: tables/imagenet.tex
\begin{table}[t]
  \centering
  \resizebox{0.95\linewidth}{!}{
\begin{tabular}{c|c|c|c}
\hline
\textbf{Method} & \textbf{Backbone}  & \textbf{Training Data} & \textbf{Acc.(\%)}                   \\ \hline
DeiT-Ti\citep{touvron2021training}   & ViT-Tiny  & IN1K   & 74.5                                   \\
SSTA-Ti\citep{pmlr-v162-wu22c}    & ViT-Tiny  & IN1K   & 75.2 \\
DMAE-Ti\citep{bai2023masked} & ViT-Tiny  &       IN1K        & 70.0 \\
MAE-Ti\citep{he2022masked}    & ViT-Tiny  & IN1K   & 75.2\\
\rowcolor{gray} SAMI-Ti (ours)  & ViT-Tiny  & SA1B (11M) + IN1K   & \textbf{76.8}                                \\ \hline
DeiT-S\citep{touvron2021training}  & ViT-Small &    IN1K & 81.2                                 \\
SSTA-S\citep{pmlr-v162-wu22c}    & ViT-Small  & IN1K   & 81.4 \\
DMAE-S\citep{bai2023masked} & ViT-Small  &       IN1K        & 79.3 \\
MAE-S\citep{he2022masked} & ViT-Small &      IN1K      & 81.5   \\
BEiT-S\citep{bao2021beit} & ViT-Small &      D250M+IN22K+IN1K         & 81.7                                     \\
CAE-S\citep{chen2023context} & ViT-Small &     D250M+IN1K        & 82.0                                  \\
DINO-S\citep{caron2021emerging} & ViT-Small &      IN1K         & 82.0  \\
iBOT-S\citep{zhou2021ibot}& ViT-Small &       IN22K+1N1K        & 82.3  \\
\rowcolor{gray} SAMI-S (ours)  & ViT-Small &     SA1B (11M) + IN1K           & \textbf{82.7}                                 \\ \hline
DeiT-B\citep{touvron2021training}  & ViT-Base  &      IN1K         & 83.8                                   \\
DMAE-B\citep{bai2023masked} & ViT-Base  &       IN1K        & 84.0                                    \\
BootMAE\citep{dong2022bootstrapped} & ViT-Base & IN1K & 84.2 \\
MAE-B\citep{he2022masked} & ViT-Base  &       IN1K        & 83.6                                    \\
BEiT-B\citep{bao2021beit} & ViT-Base  &       D250M+IN22K+IN1K        & 83.7                                 \\
CAE-B\citep{chen2023context} & ViT-Base  &       D250M+IN1K        & 83.9                                 \\
DINO-B\citep{caron2021emerging} & ViT-Base  &        IN1K       & 82.8                                \\
iBOT-B\citep{zhou2021ibot}& ViT-Base  &       IN22K+1N1K        & 84.4                                  \\
\rowcolor{gray} SAMI-B (ours) & ViT-Base  &    SA1B (11M) + IN1K           & \textbf{84.8}                                  \\ \hline
\end{tabular}}
\caption{Image classification results on ImageNet-1K. IN is short for ImageNet.}
\label{tab:imagenet}
\end{table}

%% file: tables/coco.tex
\begin{table}[t]
\centering
  \resizebox{0.75\linewidth}{!}{
\begin{tabular}{c|c|cc}
\hline
\textbf{Method} & \textbf{Backbone}  & \textbf{AP}$^{\text{bbox}}$                    & \textbf{AP}$^{\text{mask}}$                   \\ \hline
MAE-Ti\citep{he2022masked}    & ViT-Tiny  & 37.9                     & 34.9                     \\
\rowcolor{gray} SAMI-Ti(ours)   & ViT-Tiny  & \textbf{44.7}                     & \textbf{40.0}                     \\ \hline
MAE-S\citep{he2022masked}  & ViT-Small & 45.3 & 40.8 \\
DeiT-S\citep{touvron2021training} & ViT-Small & 47.2                     & 41.9                     \\
DINO-S\citep{caron2021emerging} & ViT-Small & 49.1 & 43.3 \\
iBOT-S\citep{zhou2021ibot} & ViT-Small & 49.7 & 44.0 \\
\rowcolor{gray} SAMI-S (ours)   & ViT-Small & \textbf{49.8}                     & \textbf{44.2}                     \\ \hline
MAE-B\citep{he2022masked}  & ViT-Base  & 51.6                     & 45.9                     \\
\rowcolor{gray} SAMI-B (ours) & ViT-Base  & \textbf{52.5}                     & \textbf{46.5}                     \\ \hline
\end{tabular}}
\caption{Object detection and instance segmentation results on the
MS COCO using ViTDet.}
\label{tab:coco}
\end{table}

%% file: tables/ade20k.tex
\begin{table}[t]
\centering
  \resizebox{0.65\linewidth}{!}{
\begin{tabular}{c|c|c}
\hline
\textbf{Method} & \textbf{Backbone}  & \textbf{mIOU}                     \\ \hline
MAE-Ti\citep{he2022masked}    & ViT-Tiny  & 39.0                     \\
\rowcolor{gray} SAMI-Ti(ours)   & ViT-Tiny  & \textbf{42.7}                     \\ \hline
MAE-S\citep{he2022masked}  & ViT-Small & 44.1 \\
\rowcolor{gray} SAMI-S (ours)   & ViT-Small & \textbf{48.8}                     \\ \hline
MAE-B\citep{he2022masked}  & ViT-Base  & 49.3                     \\
\rowcolor{gray} SAMI-B (ours) & ViT-Base  & \textbf{51.8}                     \\ \hline
\end{tabular}}
\caption{Semantic segmentation results on the ADE20K dataset using Mask2former. The input resolution is $512\times 512$.}
\label{tab:ade20k}
\end{table}

%% file: tables/sam_zero_shot_point.tex
\begin{table}[t]
\centering
  \resizebox{0.95\linewidth}{!}{
\begin{tabular}{c|ccc|ccc}
\hline
\multirow{2}{*}{\textbf{Method}} & \multicolumn{3}{c|}{\textbf{COCO}}                     & \multicolumn{3}{c}{\textbf{LVIS}}                      \\ \cline{2-7} 
                        & \multicolumn{1}{c|}{box}  & 1 click & 3 click & \multicolumn{1}{c|}{box}  & 1 click & 3 click \\ \hline
SAM\citep{kirillov2023segment}                     & \multicolumn{1}{c|}{78.4} & 55.6    & 74.1    & \multicolumn{1}{c|}{78.9} & 59.8    & 75.2    \\
MobileSAM\citep{zhang2023faster}               & \multicolumn{1}{c|}{74.2}   &  \multicolumn{1}{c}{43.7}   & 59.7    & \multicolumn{1}{c|}{73.8}   &     \multicolumn{1}{c}{51.0}    &    \multicolumn{1}{c}{54.4}     \\ 
SAM-MAE-Ti\citep{kirillov2023segment}               & \multicolumn{1}{c|}{74.7} & 43.3    & 65.8    & \multicolumn{1}{c|}{73.8} & 50.6    & 65.3    \\ \hline
\rowcolor{gray} EfficientSAM-Ti (ours)         & \multicolumn{1}{c|}{75.7} & 45.5    & 67.2    & \multicolumn{1}{c|}{74.3} & 52.7    & 66.8    \\
\rowcolor{gray} EfficientSAM-S (ours)         & \multicolumn{1}{c|}{76.9} & 50.0    & 69.8    & \multicolumn{1}{c|}{75.4} & 56.2    & 68.7    \\ \hline
\end{tabular}}
\caption{Zero-shot single point valid mask evaluation results on COCO and LVIS. Following SAM\citep{kirillov2023segment}, we uniformly sample random points within ground truth mask for click, and compute the tightest bounding box corresponding ground truth mask for box. SAM-MAE-Ti denotes SAM with pretrained MAE-Ti image encoder.}
\label{tab:sam0point}
\end{table}

%% file: tables/sam_zero_shot_inseg.tex
\begin{table}[t]
\centering
  \resizebox{0.95\linewidth}{!}{
\begin{tabular}{c|cccc|cccc}
\hline
\multirow{2}{*}{\textbf{Method}} & \multicolumn{4}{c|}{\textbf{COCO}} & \multicolumn{4}{c}{\textbf{LVIS}}  \\ \cline{2-9} 
                        & AP   & AP$^S$  & AP$^M$  & AP$^L$  & AP   & AP$^S$  & AP$^M$  & AP$^L$  \\ \hline
ViTDet-H\citep{li2022exploring}                & 51.0 & 32.0 & 54.3 & 68.9 & 46.6 & 35.0 & 58.0 & 66.3 \\ \hline
SAM\citep{kirillov2023segment}                     & 46.5 & 30.8 & 51.0 & 61.7 & 44.7 & 32.5 & 57.6 & 65.5 \\
MobileSAM\citep{zhang2023faster}               & 38.7    & 23.7   & 42.2    & 54.3    & 34.4    & 23.8    & 44.9    & 53.7    \\
FastSAM\citep{zhao2023fast}                 & 37.9 & 23.9 & 43.4 & 50.0 & 34.5 & 24.6 & 46.2 & 50.8 \\ \hline
\rowcolor{gray} EfficientSAM-Ti (ours)        & 42.3 & 26.7 & 46.2 & 57.4 & 39.9 & 28.9 & 51.0 & 59.9 \\
\rowcolor{gray} EfficientSAM-S (ours)        & 44.4 & 28.4 & 48.3 & 60.1 & 42.3    & 30.8    & 54.0    & 62.3    \\ \hline
\end{tabular}}
\caption{Zero-shot instance segmentation results on COCO/LVIS. ViTDet boxes are prompted to perform zero-shot segmentation. }
\label{tab:sam0inseg}
\end{table}

%% file: tables/loss.tex
\begin{table}[t]
\centering
  \resizebox{0.6\linewidth}{!}{
\begin{tabular}{c|c|c}
\hline
\textbf{Method}  & \textbf{Loss}       & \textbf{Top-1 Acc.(\%)} \\ \hline
SAMI-Ti & 1 - Cosine & 76.1       \\ \hline
\rowcolor{gray} SAMI-Ti &  MSE        & \textbf{76.8}       \\ \hline
SAMI-S  & 1 - Cosine & 82.3       \\ \hline
\rowcolor{gray} SAMI-S  &  MSE        & \textbf{82.7}       \\ \hline
\end{tabular}}
\caption{Ablation study on training loss of SAMI. MSE loss gives better classification results on ImageNet-1K.}
\label{tab:loss}
\end{table}

%% file: tables/mask_ratio.tex
\begin{table}[t]
\centering
  \resizebox{0.6\linewidth}{!}{
\begin{tabular}{c|ccc}
\hline
Mask Ratio & 50\% & 75\% & 85\% \\ \hline
Top-1 Acc.(\%) &   84.6  & \textbf{84.8} &   84.7   \\ \hline
\end{tabular}}
\caption{Ablation on the mask ratio for SAMI-B on ImageNet-1K.}
\label{tab:maskratio}
\end{table}

%% file: sec/5_conclusion.tex
\section{Conclusion}
We proposed a masked image pretraining approach, SAMI, to explore the potential of ViTs under the guidance of SAM foundation model. SAMI improves masked image pretraining by reconstructing the latent features from SAM image encoder to transfer knowledge from vision foundation model to ViTs. Extensive experiments on image classification, object detection and instance segmentation, semantic segmentation, and the segment anything task consistently validate SAMI's advantages. We also demonstrate that SAMI helps build efficient SAMs with pretrained light-weight encoders. Our preliminary work suggests that SAMI has potential applications beyond efficient segment anything task.

%% file: sec/6_suppl.tex
\clearpage
\setcounter{page}{1}
\maketitlesupplementary

In this supplementary material, we provide more results to demonstrate the instance segmentation capabilities of our efficient SAM model.
\section{Efficiency Evaluation}
\input{tables/efficiency}
Throughput and number of parameters of our models are recorded in \cref{tab:efficiency}. We measure throughput (images per second) on a single NVIDIA A100 with one box prompt. The input image resolution is $1024\times 1024$. 

\section{Qualitative Evaluation}
To study how well our model is able to produce segmentation masks based on the prompt, we use our model to perform prompt-based instance segmentation including point-based and box-based prompt prompt segmentation. We also take our model to perform segment everything and salient instance segmentation without manually creating point and box prompt. 

\begin{figure}
    \centering
    \begin{overpic}[width=1.0\linewidth]{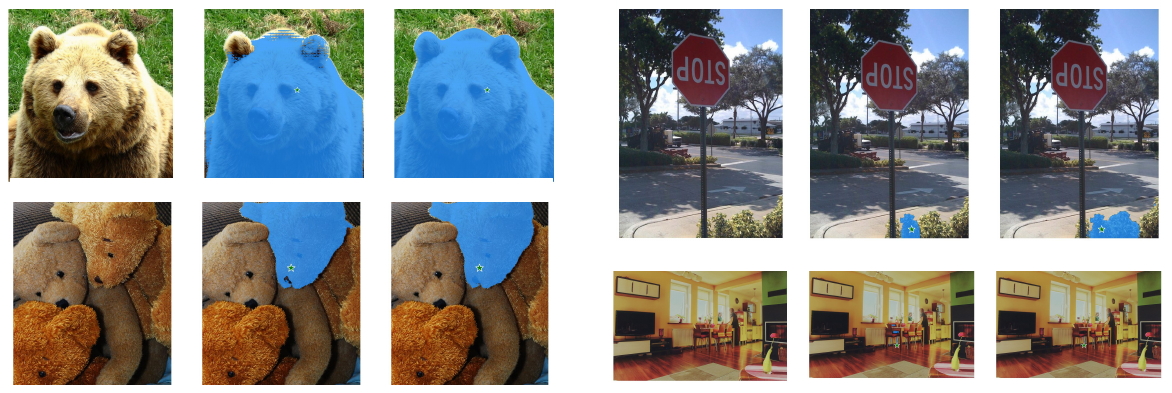}
\put (1,35) {\scriptsize{Input Image}}
\put (19,35) {\scriptsize{SAM\citep{kirillov2023segment}}}
\put (32,35) {\scriptsize{EfficientSAM}}
\put (53,35) {\scriptsize{Input Image}}
\put (71,35) {\scriptsize{SAM\citep{kirillov2023segment}}}
\put (83,35) {\scriptsize{EfficientSAM}}
\end{overpic}
\vspace{-20pt}
    \caption{Visualization results on point-prompt input.}
    \label{fig:app_point}
\end{figure}

\begin{figure}
\vspace{2pt}
    \centering
    \begin{overpic}[width=1.0\linewidth]{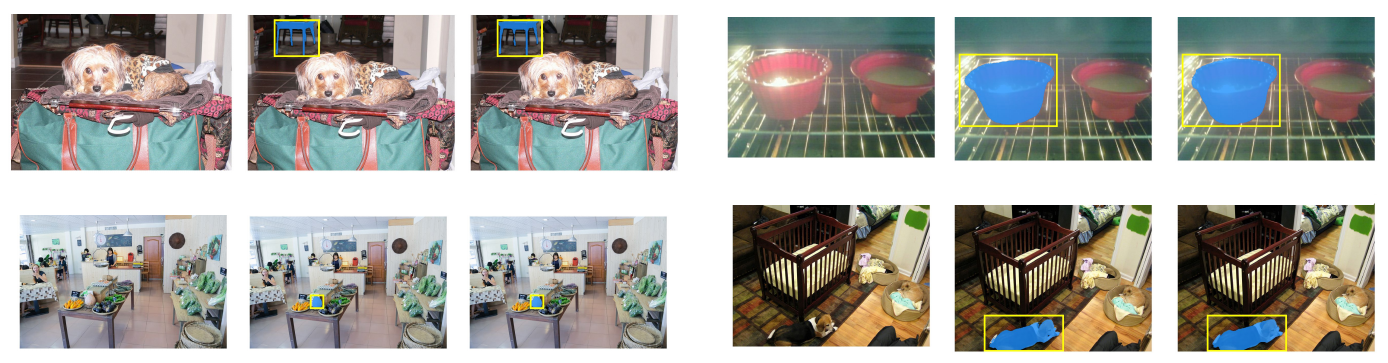}
\put (2,27) {\scriptsize{Input Image}}
\put (20,27) {\scriptsize{SAM\citep{kirillov2023segment}}}
\put (34,27) {\scriptsize{EfficientSAM}}
\put (53,27) {\scriptsize{Input Image}}
\put (71,27) {\scriptsize{SAM\citep{kirillov2023segment}}}
\put (84,27) {\scriptsize{EfficientSAM}}
\end{overpic}
\vspace{-20pt}
    \caption{Visualization results on box-prompt input.}
    \label{fig:app_box}
\end{figure}

\begin{figure}
\vspace{2pt}
    \centering
    \begin{overpic}[width=1.0\linewidth]{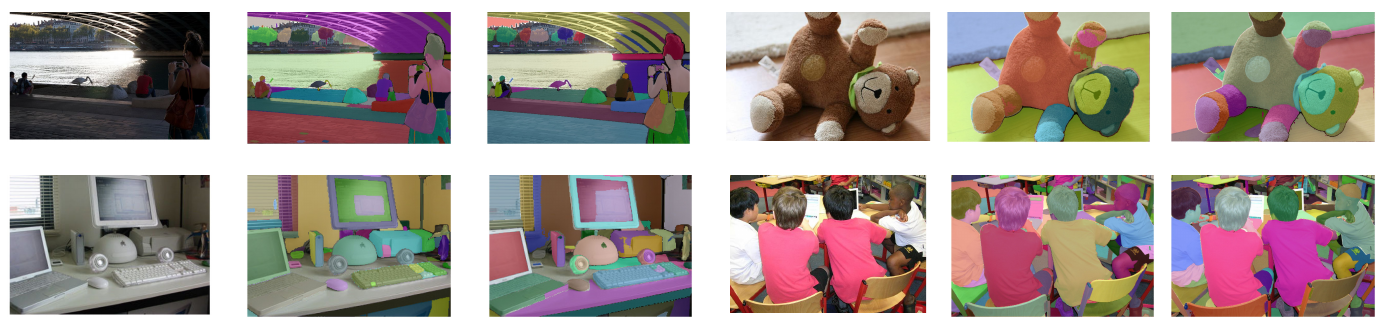}
\put (1,24) {\scriptsize{Input Image}}
\put (19.5,24) {\scriptsize{SAM\citep{kirillov2023segment}}}
\put (34,24) {\scriptsize{EfficientSAM}}
\put (53,24) {\scriptsize{Input Image}}
\put (71,24) {\scriptsize{SAM\citep{kirillov2023segment}}}
\put (84,24) {\scriptsize{EfficientSAM}}
\end{overpic}
\vspace{-20pt}
    \caption{Visualization results on segment everything.}
    \label{fig:app_every}
\end{figure}

\begin{figure}
\vspace{2pt}
    \centering
    \begin{overpic}[width=1.0\linewidth]{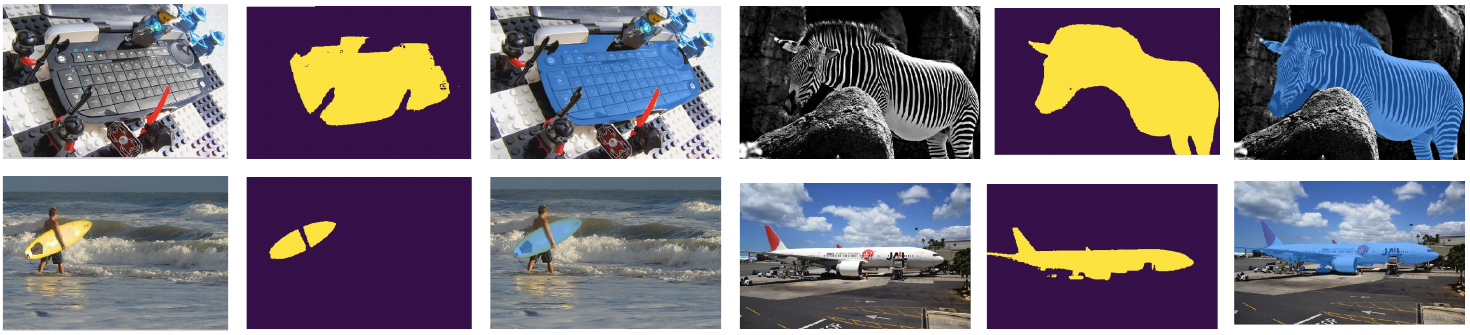}
\put (1,24) {\scriptsize{Input Image}}
\put (20.5,24) {\scriptsize{Mask}}
\put (33,24) {\scriptsize{Segmentation}}
\put (52,24) {\scriptsize{Input Image}}
\put (71.5,24) {\scriptsize{Mask}}
\put (84,24) {\scriptsize{Segmentation}}
\end{overpic}
\vspace{-20pt}
    \caption{Saliency-based automatic instance segmentation results.}
    \label{fig:app_sal}
\end{figure}

\begin{figure}
\vspace{5pt}
    \centering
    \begin{overpic}[width=0.8\linewidth]{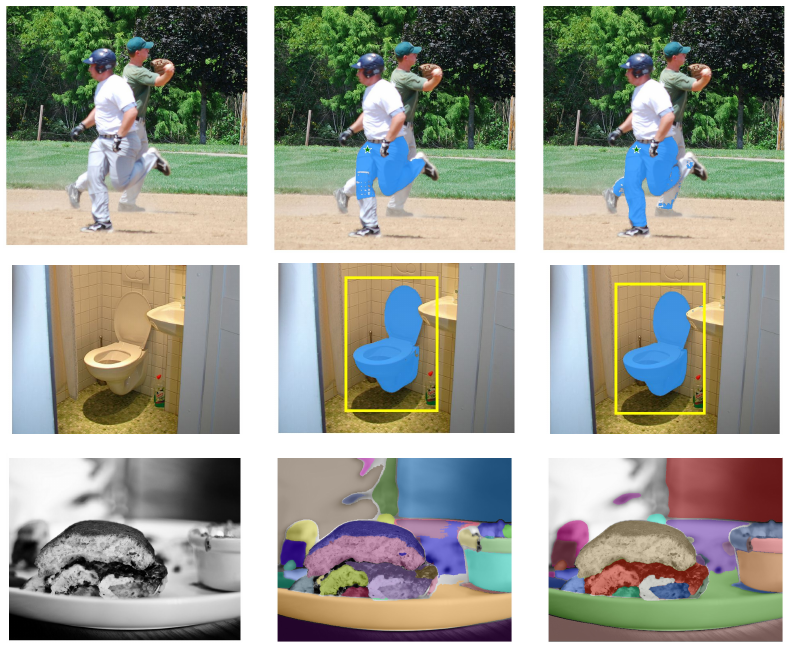}
\put (5,85) {Input Image}
\put (42.5,85) {SAM\citep{kirillov2023segment}}
\put (72.5,85) {EfficientSAM}
\end{overpic}
\vspace{-12pt}
    \caption{Segmentation results with noise. Our model may sometimes provide masks with noise, which can also be observed in the results of SAM\citep{kirillov2023segment}.}
    \label{fig:app_noise}
\end{figure}

For each task, we share 4 examples for showing the instance segmentation capabilities of our model. These results provide direct evidence for the competing instance segmentation capabilities of our EfficientSAM with different prompts. For example, in the case of point-prompt instance segmentation, our model is able to give reasonable instance segmentation results (see \cref{fig:app_point}). In the case of box-prompt instance segmentation, our model also generates expected object segmentation (see \cref{fig:app_box}). In the case of segment everything, our model provides decent segmentation performance (see \cref{fig:app_every}). In the case of salient instance segmentation, our model has the ability of generating mask and gives automatic instance segmentation without manually creating points or boxes (see \cref{fig:app_sal}). But we still need to note that our model may sometimes produce noisy segmentation, shown in \cref{fig:app_noise}.
% when the background color is close to object of interest, certain parts of the background may be included for object segmentation. 

%% file: tables/efficiency.tex
\begin{table}[t]
\centering
  \resizebox{0.8\linewidth}{!}{
\begin{tabular}{c|c|c}
\hline
\textbf{Method}          & \begin{tabular}[c]{@{}c@{}}\textbf{Params}\\ \textbf{(M)}\end{tabular} & \begin{tabular}[c]{@{}c@{}}\textbf{Throughput}\\ \textbf{(Images/Second)}\end{tabular} \\ \hline
SAM\citep{kirillov2023segment}             & 636    &   2    \\ \hline
\rowcolor{gray} EfficientSAM-Ti (ours) & 10     &   54      \\
\rowcolor{gray} EfficientSAM-S (ours)  & 25     &   47      \\ \hline
\end{tabular}
}
\vspace{-5pt}
\caption{Inference efficiency results. All models are prompted with one ViTDet\citep{li2022exploring} box for benchmarking the speed (throughput) of instance segmentation on a single NVIDIA A100.}
\label{tab:efficiency}
\end{table}

%% file: main.bbl
\begin{thebibliography}{73}
\providecommand{\natexlab}[1]{#1}
\providecommand{\url}[1]{\texttt{#1}}
\expandafter\ifx\csname urlstyle\endcsname\relax
  \providecommand{\doi}[1]{doi: #1}\else
  \providecommand{\doi}{doi: \begingroup \urlstyle{rm}\Url}\fi

\bibitem[Arbelaez et~al.(2010)Arbelaez, Maire, Fowlkes, and Malik]{arbelaez2010contour}
Pablo Arbelaez, Michael Maire, Charless Fowlkes, and Jitendra Malik.
\newblock Contour detection and hierarchical image segmentation.
\newblock \emph{IEEE transactions on pattern analysis and machine intelligence}, 33\penalty0 (5):\penalty0 898--916, 2010.

\bibitem[Bai et~al.(2023)Bai, Wang, Xiao, Wei, Wang, Yuille, Zhou, and Xie]{bai2023masked}
Yutong Bai, Zeyu Wang, Junfei Xiao, Chen Wei, Huiyu Wang, Alan~L Yuille, Yuyin Zhou, and Cihang Xie.
\newblock Masked autoencoders enable efficient knowledge distillers.
\newblock In \emph{Proceedings of the IEEE/CVF Conference on Computer Vision and Pattern Recognition}, pages 24256--24265, 2023.

\bibitem[Bao et~al.(2021)Bao, Dong, Piao, and Wei]{bao2021beit}
Hangbo Bao, Li Dong, Songhao Piao, and Furu Wei.
\newblock Beit: Bert pre-training of image transformers.
\newblock \emph{arXiv preprint arXiv:2106.08254}, 2021.

\bibitem[Borji et~al.(2019)Borji, Cheng, Hou, Jiang, and Li]{borji2019salient}
Ali Borji, Ming-Ming Cheng, Qibin Hou, Huaizu Jiang, and Jia Li.
\newblock Salient object detection: A survey.
\newblock \emph{Computational visual media}, 5:\penalty0 117--150, 2019.

\bibitem[Carion et~al.(2020)Carion, Massa, Synnaeve, Usunier, Kirillov, and Zagoruyko]{carion2020end}
Nicolas Carion, Francisco Massa, Gabriel Synnaeve, Nicolas Usunier, Alexander Kirillov, and Sergey Zagoruyko.
\newblock End-to-end object detection with transformers.
\newblock In \emph{European conference on computer vision}, pages 213--229. Springer, 2020.

\bibitem[Caron et~al.(2021)Caron, Touvron, Misra, J{\'e}gou, Mairal, Bojanowski, and Joulin]{caron2021emerging}
Mathilde Caron, Hugo Touvron, Ishan Misra, Herv{\'e} J{\'e}gou, Julien Mairal, Piotr Bojanowski, and Armand Joulin.
\newblock Emerging properties in self-supervised vision transformers.
\newblock In \emph{Proceedings of the IEEE/CVF international conference on computer vision}, pages 9650--9660, 2021.

\bibitem[Cen et~al.(2023)Cen, Wu, Wang, Li, Yang, Pei, Kong, Liu, and Chen]{cen2023sad}
Jun Cen, Yizheng Wu, Kewei Wang, Xingyi Li, Jingkang Yang, Yixuan Pei, Lingdong Kong, Ziwei Liu, and Qifeng Chen.
\newblock Sad: Segment any rgbd.
\newblock \emph{arXiv preprint arXiv:2305.14207}, 2023.

\bibitem[Chen et~al.(2023{\natexlab{a}})Chen, Yang, and Zhang]{chen2023semantic}
Jiaqi Chen, Zeyu Yang, and Li Zhang.
\newblock Semantic segment anything.
\newblock \url{https://github.com/fudan-zvg/Semantic-Segment-Anything}, 2023{\natexlab{a}}.

\bibitem[Chen et~al.(2020)Chen, Kornblith, Norouzi, and Hinton]{chen2020simple}
Ting Chen, Simon Kornblith, Mohammad Norouzi, and Geoffrey Hinton.
\newblock A simple framework for contrastive learning of visual representations.
\newblock In \emph{International conference on machine learning}, pages 1597--1607. PMLR, 2020.

\bibitem[Chen et~al.(2023{\natexlab{b}})Chen, Zhu, Deng, Cao, Wang, Zhang, Li, Sun, Zang, and Mao]{chen2023sam}
Tianrun Chen, Lanyun Zhu, Chaotao Deng, Runlong Cao, Yan Wang, Shangzhan Zhang, Zejian Li, Lingyun Sun, Ying Zang, and Papa Mao.
\newblock Sam-adapter: Adapting segment anything in underperformed scenes.
\newblock In \emph{Proceedings of the IEEE/CVF International Conference on Computer Vision}, pages 3367--3375, 2023{\natexlab{b}}.

\bibitem[Chen and He(2021)]{chen2021exploring}
Xinlei Chen and Kaiming He.
\newblock Exploring simple siamese representation learning.
\newblock In \emph{Proceedings of the IEEE/CVF conference on computer vision and pattern recognition}, pages 15750--15758, 2021.

\bibitem[Chen et~al.(2023{\natexlab{c}})Chen, Ding, Wang, Xin, Mo, Wang, Han, Luo, Zeng, and Wang]{chen2023context}
Xiaokang Chen, Mingyu Ding, Xiaodi Wang, Ying Xin, Shentong Mo, Yunhao Wang, Shumin Han, Ping Luo, Gang Zeng, and Jingdong Wang.
\newblock Context autoencoder for self-supervised representation learning.
\newblock \emph{International Journal of Computer Vision}, pages 1--16, 2023{\natexlab{c}}.

\bibitem[Cheng et~al.(2022)Cheng, Misra, Schwing, Kirillov, and Girdhar]{cheng2022masked}
Bowen Cheng, Ishan Misra, Alexander~G Schwing, Alexander Kirillov, and Rohit Girdhar.
\newblock Masked-attention mask transformer for universal image segmentation.
\newblock In \emph{Proceedings of the IEEE/CVF conference on computer vision and pattern recognition}, pages 1290--1299, 2022.

\bibitem[Cheng et~al.(2023)Cheng, Li, Xu, Li, Yang, Wang, and Yang]{cheng2023segment}
Yangming Cheng, Liulei Li, Yuanyou Xu, Xiaodi Li, Zongxin Yang, Wenguan Wang, and Yi Yang.
\newblock Segment and track anything.
\newblock \emph{arXiv preprint arXiv:2305.06558}, 2023.

\bibitem[Cubuk et~al.(2020)Cubuk, Zoph, Shlens, and Le]{cubuk2020randaugment}
Ekin~D Cubuk, Barret Zoph, Jonathon Shlens, and Quoc~V Le.
\newblock Randaugment: Practical automated data augmentation with a reduced search space.
\newblock In \emph{Proceedings of the IEEE/CVF conference on computer vision and pattern recognition workshops}, pages 702--703, 2020.

\bibitem[Deng et~al.(2009)Deng, Dong, Socher, Li, Li, and Fei-Fei]{deng2009imagenet}
Jia Deng, Wei Dong, Richard Socher, Li-Jia Li, Kai Li, and Li Fei-Fei.
\newblock Imagenet: A large-scale hierarchical image database.
\newblock In \emph{2009 IEEE conference on computer vision and pattern recognition}, pages 248--255. Ieee, 2009.

\bibitem[Deng et~al.(2023)Deng, Cui, Liu, Yao, Remedios, Bao, Landman, Wheless, Coburn, Wilson, et~al.]{deng2023segment}
Ruining Deng, Can Cui, Quan Liu, Tianyuan Yao, Lucas~W Remedios, Shunxing Bao, Bennett~A Landman, Lee~E Wheless, Lori~A Coburn, Keith~T Wilson, et~al.
\newblock Segment anything model (sam) for digital pathology: Assess zero-shot segmentation on whole slide imaging.
\newblock \emph{arXiv preprint arXiv:2304.04155}, 2023.

\bibitem[Dong et~al.(2022)Dong, Bao, Zhang, Chen, Zhang, Yuan, Chen, Wen, and Yu]{dong2022bootstrapped}
Xiaoyi Dong, Jianmin Bao, Ting Zhang, Dongdong Chen, Weiming Zhang, Lu Yuan, Dong Chen, Fang Wen, and Nenghai Yu.
\newblock Bootstrapped masked autoencoders for vision bert pretraining.
\newblock In \emph{European Conference on Computer Vision}, pages 247--264. Springer, 2022.

\bibitem[Dosovitskiy et~al.(2020)Dosovitskiy, Beyer, Kolesnikov, Weissenborn, Zhai, Unterthiner, Dehghani, Minderer, Heigold, Gelly, et~al.]{dosovitskiy2020image}
Alexey Dosovitskiy, Lucas Beyer, Alexander Kolesnikov, Dirk Weissenborn, Xiaohua Zhai, Thomas Unterthiner, Mostafa Dehghani, Matthias Minderer, Georg Heigold, Sylvain Gelly, et~al.
\newblock An image is worth 16x16 words: Transformers for image recognition at scale.
\newblock \emph{arXiv preprint arXiv:2010.11929}, 2020.

\bibitem[Fan et~al.(2021)Fan, Xiong, Mangalam, Li, Yan, Malik, and Feichtenhofer]{fan2021multiscale}
Haoqi Fan, Bo Xiong, Karttikeya Mangalam, Yanghao Li, Zhicheng Yan, Jitendra Malik, and Christoph Feichtenhofer.
\newblock Multiscale vision transformers.
\newblock In \emph{Proceedings of the IEEE/CVF international conference on computer vision}, pages 6824--6835, 2021.

\bibitem[Gao et~al.(2023)Gao, Lin, Xie, Zhou, Cheng, and Yan]{gao2023editanything}
Shanghua Gao, Zhijie Lin, Xingyu Xie, Pan Zhou, Ming-Ming Cheng, and Shuicheng Yan.
\newblock Editanything: Empowering unparalleled flexibility in image editing and generation.
\newblock In \emph{Proceedings of the 31st ACM International Conference on Multimedia}, pages 9414--9416, 2023.

\bibitem[Graham et~al.(2021)Graham, El-Nouby, Touvron, Stock, Joulin, J{\'e}gou, and Douze]{graham2021levit}
Benjamin Graham, Alaaeldin El-Nouby, Hugo Touvron, Pierre Stock, Armand Joulin, Herv{\'e} J{\'e}gou, and Matthijs Douze.
\newblock Levit: a vision transformer in convnet's clothing for faster inference.
\newblock In \emph{Proceedings of the IEEE/CVF international conference on computer vision}, pages 12259--12269, 2021.

\bibitem[Gupta et~al.(2019)Gupta, Dollár, and Girshick]{gupta2019lvis}
Agrim Gupta, Piotr Dollár, and Ross Girshick.
\newblock Lvis: A dataset for large vocabulary instance segmentation.
\newblock In \emph{Proceedings of the IEEE/CVF Conference on Computer Vision and Pattern Recognition}, 2019.

\bibitem[Han et~al.(2023)Han, Zhang, Qiao, Qamar, Jung, Lee, Bae, and Hong]{han2023segment}
Dongsheng Han, Chaoning Zhang, Yu Qiao, Maryam Qamar, Yuna Jung, SeungKyu Lee, Sung-Ho Bae, and Choong~Seon Hong.
\newblock Segment anything model (sam) meets glass: Mirror and transparent objects cannot be easily detected.
\newblock \emph{arXiv preprint arXiv:2305.00278}, 2023.

\bibitem[He et~al.(2017)He, Gkioxari, Doll{\'a}r, and Girshick]{he2017mask}
Kaiming He, Georgia Gkioxari, Piotr Doll{\'a}r, and Ross Girshick.
\newblock Mask r-cnn.
\newblock In \emph{Proceedings of the IEEE international conference on computer vision}, pages 2961--2969, 2017.

\bibitem[He et~al.(2022)He, Chen, Xie, Li, Doll{\'a}r, and Girshick]{he2022masked}
Kaiming He, Xinlei Chen, Saining Xie, Yanghao Li, Piotr Doll{\'a}r, and Ross Girshick.
\newblock Masked autoencoders are scalable vision learners.
\newblock In \emph{Proceedings of the IEEE/CVF conference on computer vision and pattern recognition}, pages 16000--16009, 2022.

\bibitem[Hinton et~al.(2015)Hinton, Vinyals, and Dean]{hinton2015distilling}
Geoffrey Hinton, Oriol Vinyals, and Jeff Dean.
\newblock Distilling the knowledge in a neural network.
\newblock \emph{arXiv preprint arXiv:1503.02531}, 2015.

\bibitem[Hou et~al.(2022)Hou, Sun, Chen, Xie, and Kung]{hou2022milan}
Zejiang Hou, Fei Sun, Yen-Kuang Chen, Yuan Xie, and Sun-Yuan Kung.
\newblock Milan: Masked image pretraining on language assisted representation.
\newblock \emph{arXiv preprint arXiv:2208.06049}, 2022.

\bibitem[Jiang and Holz(2023)]{jiang2023restore}
Jiaxi Jiang and Christian Holz.
\newblock Restore anything pipeline: Segment anything meets image restoration.
\newblock \emph{arXiv preprint arXiv:2305.13093}, 2023.

\bibitem[Jocher et~al.(2023)Jocher, Chaurasia, and Qiu]{yolov8}
Glenn Jocher, Ayush Chaurasia, and Jing Qiu.
\newblock Yolo by ultralytics.
\newblock \url{https://github.com/ultralytics/ultralytics}, 2023.

\bibitem[Kirillov et~al.(2023)Kirillov, Mintun, Ravi, Mao, Rolland, Gustafson, Xiao, Whitehead, Berg, Lo, et~al.]{kirillov2023segment}
Alexander Kirillov, Eric Mintun, Nikhila Ravi, Hanzi Mao, Chloe Rolland, Laura Gustafson, Tete Xiao, Spencer Whitehead, Alexander~C Berg, Wan-Yen Lo, et~al.
\newblock Segment anything.
\newblock \emph{arXiv preprint arXiv:2304.02643}, 2023.

\bibitem[Koonce and Koonce(2021)]{koonce2021mobilenetv3}
Brett Koonce and Brett Koonce.
\newblock Mobilenetv3.
\newblock \emph{Convolutional Neural Networks with Swift for Tensorflow: Image Recognition and Dataset Categorization}, pages 125--144, 2021.

\bibitem[Li et~al.(2022{\natexlab{a}})Li, Xia, Li, Li, Wang, Xiao, Wang, Zheng, and Pan]{li2022next}
Jiashi Li, Xin Xia, Wei Li, Huixia Li, Xing Wang, Xuefeng Xiao, Rui Wang, Min Zheng, and Xin Pan.
\newblock Next-vit: Next generation vision transformer for efficient deployment in realistic industrial scenarios.
\newblock \emph{arXiv preprint arXiv:2207.05501}, 2022{\natexlab{a}}.

\bibitem[Li et~al.(2022{\natexlab{b}})Li, Mao, Girshick, and He]{li2022exploring}
Yanghao Li, Hanzi Mao, Ross Girshick, and Kaiming He.
\newblock Exploring plain vision transformer backbones for object detection.
\newblock In \emph{European Conference on Computer Vision}, pages 280--296. Springer, 2022{\natexlab{b}}.

\bibitem[Li et~al.(2022{\natexlab{c}})Li, Yuan, Wen, Hu, Evangelidis, Tulyakov, Wang, and Ren]{li2022efficientformer}
Yanyu Li, Geng Yuan, Yang Wen, Ju Hu, Georgios Evangelidis, Sergey Tulyakov, Yanzhi Wang, and Jian Ren.
\newblock Efficientformer: Vision transformers at mobilenet speed.
\newblock \emph{Advances in Neural Information Processing Systems}, 35:\penalty0 12934--12949, 2022{\natexlab{c}}.

\bibitem[Lin et~al.(2014)Lin, Maire, Belongie, Hays, Perona, Ramanan, Doll{\'a}r, and Zitnick]{lin2014microsoft}
Tsung-Yi Lin, Michael Maire, Serge Belongie, James Hays, Pietro Perona, Deva Ramanan, Piotr Doll{\'a}r, and C~Lawrence Zitnick.
\newblock Microsoft coco: Common objects in context.
\newblock In \emph{European conference on computer vision}, pages 740--755. Springer, 2014.

\bibitem[Liu et~al.(2023{\natexlab{a}})Liu, Zhang, Peng, Zheng, Cao, Chen, Yang, and Stiefelhagen]{liu2023open}
Ruiping Liu, Jiaming Zhang, Kunyu Peng, Junwei Zheng, Ke Cao, Yufan Chen, Kailun Yang, and Rainer Stiefelhagen.
\newblock Open scene understanding: Grounded situation recognition meets segment anything for helping people with visual impairments.
\newblock In \emph{Proceedings of the IEEE/CVF International Conference on Computer Vision}, pages 1857--1867, 2023{\natexlab{a}}.

\bibitem[Liu et~al.(2023{\natexlab{b}})Liu, Peng, Zheng, Yang, Hu, and Yuan]{liu2023efficientvit}
Xinyu Liu, Houwen Peng, Ningxin Zheng, Yuqing Yang, Han Hu, and Yixuan Yuan.
\newblock Efficientvit: Memory efficient vision transformer with cascaded group attention.
\newblock In \emph{Proceedings of the IEEE/CVF Conference on Computer Vision and Pattern Recognition}, pages 14420--14430, 2023{\natexlab{b}}.

\bibitem[Liu et~al.(2021)Liu, Lin, Cao, Hu, Wei, Zhang, Lin, and Guo]{liu2021swin}
Ze Liu, Yutong Lin, Yue Cao, Han Hu, Yixuan Wei, Zheng Zhang, Stephen Lin, and Baining Guo.
\newblock Swin transformer: Hierarchical vision transformer using shifted windows.
\newblock In \emph{Proceedings of the IEEE/CVF international conference on computer vision}, pages 10012--10022, 2021.

\bibitem[Loshchilov and Hutter(2018)]{loshchilov2018decoupled}
Ilya Loshchilov and Frank Hutter.
\newblock Decoupled weight decay regularization.
\newblock In \emph{International Conference on Learning Representations}, 2018.

\bibitem[Ma and Wang(2023)]{ma2023segment}
Jun Ma and Bo Wang.
\newblock Segment anything in medical images.
\newblock \emph{arXiv preprint arXiv:2304.12306}, 2023.

\bibitem[Mehta and Rastegari(2021)]{mehta2021mobilevit}
Sachin Mehta and Mohammad Rastegari.
\newblock Mobilevit: Light-weight, general-purpose, and mobile-friendly vision transformer.
\newblock In \emph{International Conference on Learning Representations}, 2021.

\bibitem[Pathak et~al.(2016)Pathak, Krahenbuhl, Donahue, Darrell, and Efros]{pathak2016context}
Deepak Pathak, Philipp Krahenbuhl, Jeff Donahue, Trevor Darrell, and Alexei~A Efros.
\newblock Context encoders: Feature learning by inpainting.
\newblock In \emph{Proceedings of the IEEE conference on computer vision and pattern recognition}, pages 2536--2544, 2016.

\bibitem[Peng et~al.(2022)Peng, Dong, Bao, Ye, and Wei]{peng2022beit}
Zhiliang Peng, Li Dong, Hangbo Bao, Qixiang Ye, and Furu Wei.
\newblock Beit v2: Masked image modeling with vector-quantized visual tokenizers.
\newblock \emph{arXiv preprint arXiv:2208.06366}, 2022.

\bibitem[Qin et~al.(2020)Qin, Zhang, Huang, Dehghan, Zaiane, and Jagersand]{Qin_2020_PR}
Xuebin Qin, Zichen Zhang, Chenyang Huang, Masood Dehghan, Osmar Zaiane, and Martin Jagersand.
\newblock U2-net: Going deeper with nested u-structure for salient object detection.
\newblock page 107404, 2020.

\bibitem[Ramesh et~al.(2021)Ramesh, Pavlov, Goh, Gray, Voss, Radford, Chen, and Sutskever]{ramesh2021zero}
Aditya Ramesh, Mikhail Pavlov, Gabriel Goh, Scott Gray, Chelsea Voss, Alec Radford, Mark Chen, and Ilya Sutskever.
\newblock Zero-shot text-to-image generation.
\newblock In \emph{International Conference on Machine Learning}, pages 8821--8831. PMLR, 2021.

\bibitem[Romero et~al.(2014)Romero, Ballas, Kahou, Chassang, Gatta, and Bengio]{romero2014fitnets}
Adriana Romero, Nicolas Ballas, Samira~Ebrahimi Kahou, Antoine Chassang, Carlo Gatta, and Yoshua Bengio.
\newblock Fitnets: Hints for thin deep nets.
\newblock \emph{arXiv preprint arXiv:1412.6550}, 2014.

\bibitem[Sandler et~al.(2018)Sandler, Howard, Zhu, Zhmoginov, and Chen]{sandler2018mobilenetv2}
Mark Sandler, Andrew Howard, Menglong Zhu, Andrey Zhmoginov, and Liang-Chieh Chen.
\newblock Mobilenetv2: Inverted residuals and linear bottlenecks.
\newblock In \emph{Proceedings of the IEEE conference on computer vision and pattern recognition}, pages 4510--4520, 2018.

\bibitem[Shen et~al.(2023)Shen, Yang, and Wang]{shen2023anything}
Qiuhong Shen, Xingyi Yang, and Xinchao Wang.
\newblock Anything-3d: Towards single-view anything reconstruction in the wild.
\newblock \emph{arXiv preprint arXiv:2304.10261}, 2023.

\bibitem[Sun et~al.(2023)Sun, Ma, Yuan, and Wang]{sun2023explain}
Ao Sun, Pingchuan Ma, Yuanyuan Yuan, and Shuai Wang.
\newblock Explain any concept: Segment anything meets concept-based explanation.
\newblock \emph{arXiv preprint arXiv:2305.10289}, 2023.

\bibitem[Tang et~al.(2023)Tang, Xiao, and Li]{tang2023can}
Lv Tang, Haoke Xiao, and Bo Li.
\newblock Can sam segment anything? when sam meets camouflaged object detection.
\newblock \emph{arXiv preprint arXiv:2304.04709}, 2023.

\bibitem[Tariq et~al.(2023)Tariq, Arfeto, Zhang, and Shin]{tariq2023segment}
Shehbaz Tariq, Brian~Estadimas Arfeto, Chaoning Zhang, and Hyundong Shin.
\newblock Segment anything meets semantic communication.
\newblock \emph{arXiv preprint arXiv:2306.02094}, 2023.

\bibitem[Touvron et~al.(2021)Touvron, Cord, Douze, Massa, Sablayrolles, and J{\'e}gou]{touvron2021training}
Hugo Touvron, Matthieu Cord, Matthijs Douze, Francisco Massa, Alexandre Sablayrolles, and Herv{\'e} J{\'e}gou.
\newblock Training data-efficient image transformers \& distillation through attention.
\newblock In \emph{International conference on machine learning}, pages 10347--10357. PMLR, 2021.

\bibitem[Van~de Sande et~al.(2011)Van~de Sande, Uijlings, Gevers, and Smeulders]{van2011segmentation}
Koen~EA Van~de Sande, Jasper~RR Uijlings, Theo Gevers, and Arnold~WM Smeulders.
\newblock Segmentation as selective search for object recognition.
\newblock In \emph{2011 international conference on computer vision}, pages 1879--1886. IEEE, 2011.

\bibitem[Vaswani et~al.(2017)Vaswani, Shazeer, Parmar, Uszkoreit, Jones, Gomez, Kaiser, and Polosukhin]{vaswani2017attention}
Ashish Vaswani, Noam Shazeer, Niki Parmar, Jakob Uszkoreit, Llion Jones, Aidan~N Gomez, {\L}ukasz Kaiser, and Illia Polosukhin.
\newblock Attention is all you need.
\newblock \emph{Advances in neural information processing systems}, 30, 2017.

\bibitem[Vincent et~al.(2010)Vincent, Larochelle, Lajoie, Bengio, Manzagol, and Bottou]{vincent2010stacked}
Pascal Vincent, Hugo Larochelle, Isabelle Lajoie, Yoshua Bengio, Pierre-Antoine Manzagol, and L{\'e}on Bottou.
\newblock Stacked denoising autoencoders: Learning useful representations in a deep network with a local denoising criterion.
\newblock \emph{Journal of machine learning research}, 11\penalty0 (12), 2010.

\bibitem[Wang et~al.(2021)Wang, Zhang, Shen, Kong, and Li]{wang2021dense}
Xinlong Wang, Rufeng Zhang, Chunhua Shen, Tao Kong, and Lei Li.
\newblock Dense contrastive learning for self-supervised visual pre-training.
\newblock In \emph{Proceedings of the IEEE/CVF Conference on Computer Vision and Pattern Recognition}, pages 3024--3033, 2021.

\bibitem[Wang et~al.(2022)Wang, Zhang, Yang, and Sun]{wang2022anchor}
Yingming Wang, Xiangyu Zhang, Tong Yang, and Jian Sun.
\newblock Anchor detr: Query design for transformer-based detector.
\newblock In \emph{Proceedings of the AAAI conference on artificial intelligence}, pages 2567--2575, 2022.

\bibitem[Wei et~al.(2022)Wei, Fan, Xie, Wu, Yuille, and Feichtenhofer]{wei2022masked}
Chen Wei, Haoqi Fan, Saining Xie, Chao-Yuan Wu, Alan Yuille, and Christoph Feichtenhofer.
\newblock Masked feature prediction for self-supervised visual pre-training.
\newblock In \emph{Proceedings of the IEEE/CVF Conference on Computer Vision and Pattern Recognition}, pages 14668--14678, 2022.

\bibitem[Wu et~al.(2022{\natexlab{a}})Wu, Gao, Zhang, Lin, Xie, Sun, and Li]{pmlr-v162-wu22c}
Haiyan Wu, Yuting Gao, Yinqi Zhang, Shaohui Lin, Yuan Xie, Xing Sun, and Ke Li.
\newblock Self-supervised models are good teaching assistants for vision transformers.
\newblock In \emph{Proceedings of the 39th International Conference on Machine Learning}, pages 24031--24042. PMLR, 2022{\natexlab{a}}.

\bibitem[Wu et~al.(2022{\natexlab{b}})Wu, Zhang, Peng, Liu, Xiao, Fu, and Yuan]{wu2022tinyvit}
Kan Wu, Jinnian Zhang, Houwen Peng, Mengchen Liu, Bin Xiao, Jianlong Fu, and Lu Yuan.
\newblock Tinyvit: Fast pretraining distillation for small vision transformers.
\newblock In \emph{European Conference on Computer Vision}, pages 68--85. Springer, 2022{\natexlab{b}}.

\bibitem[Xie et~al.(2021)Xie, Lin, Zhang, Cao, Lin, and Hu]{xie2021propagate}
Zhenda Xie, Yutong Lin, Zheng Zhang, Yue Cao, Stephen Lin, and Han Hu.
\newblock Propagate yourself: Exploring pixel-level consistency for unsupervised visual representation learning.
\newblock In \emph{Proceedings of the IEEE/CVF Conference on Computer Vision and Pattern Recognition}, pages 16684--16693, 2021.

\bibitem[Xie et~al.(2022)Xie, Zhang, Cao, Lin, Bao, Yao, Dai, and Hu]{xie2022simmim}
Zhenda Xie, Zheng Zhang, Yue Cao, Yutong Lin, Jianmin Bao, Zhuliang Yao, Qi Dai, and Han Hu.
\newblock Simmim: A simple framework for masked image modeling.
\newblock In \emph{Proceedings of the IEEE/CVF Conference on Computer Vision and Pattern Recognition}, pages 9653--9663, 2022.

\bibitem[Yang et~al.(2021)Yang, Martinez, Bulat, and Tzimiropoulos]{yang2021knowledge}
Jing Yang, Brais Martinez, Adrian Bulat, and Georgios Tzimiropoulos.
\newblock Knowledge distillation via softmax regression representation learning.
\newblock In \emph{International Conference on Learning Representations}, 2021.

\bibitem[Yang et~al.(2023)Yang, Gao, Li, Gao, Wang, and Zheng]{yang2023track}
Jinyu Yang, Mingqi Gao, Zhe Li, Shang Gao, Fangjing Wang, and Feng Zheng.
\newblock Track anything: Segment anything meets videos.
\newblock \emph{arXiv preprint arXiv:2304.11968}, 2023.

\bibitem[You et~al.(2023)You, Xiong, Dai, Wu, Zhang, Fan, Vajda, and Lin]{you2023castling}
Haoran You, Yunyang Xiong, Xiaoliang Dai, Bichen Wu, Peizhao Zhang, Haoqi Fan, Peter Vajda, and Yingyan~Celine Lin.
\newblock Castling-vit: Compressing self-attention via switching towards linear-angular attention at vision transformer inference.
\newblock In \emph{Proceedings of the IEEE/CVF Conference on Computer Vision and Pattern Recognition}, pages 14431--14442, 2023.

\bibitem[Yu et~al.(2023)Yu, Feng, Feng, Liu, Jin, Zeng, and Chen]{yu2023inpaint}
Tao Yu, Runseng Feng, Ruoyu Feng, Jinming Liu, Xin Jin, Wenjun Zeng, and Zhibo Chen.
\newblock Inpaint anything: Segment anything meets image inpainting.
\newblock \emph{arXiv preprint arXiv:2304.06790}, 2023.

\bibitem[Zhang et~al.(2023{\natexlab{a}})Zhang, Han, Qiao, Kim, Bae, Lee, and Hong]{zhang2023faster}
Chaoning Zhang, Dongshen Han, Yu Qiao, Jung~Uk Kim, Sung-Ho Bae, Seungkyu Lee, and Choong~Seon Hong.
\newblock Faster segment anything: Towards lightweight sam for mobile applications.
\newblock \emph{arXiv preprint arXiv:2306.14289}, 2023{\natexlab{a}}.

\bibitem[Zhang et~al.(2023{\natexlab{b}})Zhang, Song, and Yao]{zhang2023deshadow}
Xiao~Feng Zhang, Tian~Yi Song, and Jia~Wei Yao.
\newblock Deshadow-anything: When segment anything model meets zero-shot shadow removal.
\newblock \emph{arXiv preprint arXiv:2309.11715}, 2023{\natexlab{b}}.

\bibitem[Zhao et~al.(2022)Zhao, Cui, Song, Qiu, and Liang]{zhao2022decoupled}
Borui Zhao, Quan Cui, Renjie Song, Yiyu Qiu, and Jiajun Liang.
\newblock Decoupled knowledge distillation.
\newblock In \emph{Proceedings of the IEEE/CVF Conference on computer vision and pattern recognition}, pages 11953--11962, 2022.

\bibitem[Zhao et~al.(2023)Zhao, Ding, An, Du, Yu, Li, Tang, and Wang]{zhao2023fast}
Xu Zhao, Wenchao Ding, Yongqi An, Yinglong Du, Tao Yu, Min Li, Ming Tang, and Jinqiao Wang.
\newblock Fast segment anything.
\newblock \emph{arXiv preprint arXiv:2306.12156}, 2023.

\bibitem[Zhou et~al.(2017)Zhou, Zhao, Puig, Fidler, Barriuso, and Torralba]{zhou2017scene}
Bolei Zhou, Hang Zhao, Xavier Puig, Sanja Fidler, Adela Barriuso, and Antonio Torralba.
\newblock Scene parsing through ade20k dataset.
\newblock In \emph{Proceedings of the IEEE conference on computer vision and pattern recognition}, pages 633--641, 2017.

\bibitem[Zhou et~al.(2021)Zhou, Wei, Wang, Shen, Xie, Yuille, and Kong]{zhou2021ibot}
Jinghao Zhou, Chen Wei, Huiyu Wang, Wei Shen, Cihang Xie, Alan Yuille, and Tao Kong.
\newblock ibot: Image bert pre-training with online tokenizer.
\newblock \emph{arXiv preprint arXiv:2111.07832}, 2021.

\end{thebibliography}
